%% file: main.tex
\pdfoutput=1

\documentclass[10pt,twocolumn,letterpaper]{article}

\usepackage{cvpr}              
\input{math_commands.tex}

\usepackage{graphicx}
\usepackage{amsmath}
\usepackage{amssymb}
\usepackage{booktabs}

\usepackage{array}
\usepackage{caption}
\usepackage{multirow}
\usepackage{makecell}
\usepackage{pifont}
\usepackage{siunitx}

\usepackage[accsupp]{axessibility}  

\usepackage[pagebackref,breaklinks,colorlinks]{hyperref}

\usepackage[capitalize]{cleveref}
\crefname{section}{Sec.}{Secs.}
\Crefname{section}{Section}{Sections}
\Crefname{table}{Table}{Tables}
\crefname{table}{Tab.}{Tabs.}

\def\cvprPaperID{5085} 
\def\confName{CVPR}
\def\confYear{2022}

\begin{document}

\title{Scene Graph Expansion for Semantics-Guided Image Outpainting}

\author{Chiao-An Yang$^1$, Cheng-Yo Tan$^1$, Wan-Cyuan Fan$^1$, Cheng-Fu Yang$^1$\\
Meng-Lin Wu$^2$, Yu-Chiang Frank Wang$^1$\\
$^1$ National Taiwan University, $^2$ Qualcomm Technologies, Inc. \\
{\tt\small joeyang@ntu.edu.tw, \{cy.ugo.tan, christine5200312,  joeyy5588\}@gmail.com} \\ 
{\tt \small menglinw@qti.qualcomm.com, ycwang@ntu.edu.tw}
}
\maketitle

\input{macro}

\begin{abstract}
In this paper, we address the task of semantics-guided image outpainting, which is to complete an image by generating semantically practical content. Different from most existing image outpainting works, we approach the above task by understanding and completing image semantics at the scene graph level. In particular, we propose a novel network of Scene Graph Transformer (SGT), which is designed to take node and edge features as inputs for modeling the associated structural information. To better understand and process graph-based inputs, our SGT uniquely performs feature attention at both node and edge levels. While the former views edges as relationship regularization, the latter observes the co-occurrence of nodes for guiding the attention process. We demonstrate that, given a partial input image with its layout and scene graph, our SGT can be applied for scene graph expansion and its conversion to a complete layout. Following state-of-the-art layout-to-image conversions works, the task of image outpainting can be completed with sufficient and practical semantics introduced. Extensive experiments are conducted on the datasets of MS-COCO and Visual Genome, which quantitatively and qualitatively confirm the effectiveness of our proposed SGT and outpainting frameworks.
\end{abstract}
\vspace{-2mm}

\input{1_Introduction}

\input{2_related}
\input{3_approach}

\input{4_experiment}

\input{5_conclusion}

\input{7_acknowledgement.tex}



\clearpage

{\small
\bibliographystyle{ieee_fullname}
\bibliography{main.bib}
}

\clearpage

\pagebreak



\input{6_supplementary}

\end{document}

%% file: math_commands.tex
\usepackage{amsmath,amsfonts,bm}

\def\Figref#1{Figure~\ref{#1}}

\def\eqref#1{equation~\ref{#1}}
\def\Eqref#1{Equation~\ref{#1}}

\def\1{\bm{1}}

\def\vb{{\bm{b}}}

\def\vd{{\bm{d}}}

\def\vf{{\bm{f}}}

\def\vh{{\bm{h}}}

\def\vk{{\bm{k}}}

\def\vq{{\bm{q}}}

\def\vs{{\bm{s}}}
\def\vt{{\bm{t}}}

\def\vv{{\bm{v}}}

\def\mB{{\bm{B}}}

\def\mD{{\bm{D}}}

\def\mH{{\bm{H}}}
\def\mI{{\bm{I}}}

\def\mL{{\bm{L}}}
\def\mM{{\bm{M}}}

\def\mO{{\bm{O}}}

\def\mR{{\bm{R}}}

\DeclareMathAlphabet{\mathsfit}{\encodingdefault}{\sfdefault}{m}{sl}
\SetMathAlphabet{\mathsfit}{bold}{\encodingdefault}{\sfdefault}{bx}{n}

\def\gL{{\mathcal{L}}}

\def\gS{{\mathcal{S}}}

\def\sR{{\mathbb{R}}}
\def\sS{{\mathbb{S}}}



%% file: macro.tex


\newcommand{\TODO}{{\textbf{\color{red}[TODO]\,}}}



\definecolor{gray}{rgb}{0.35,0.35,0.35}
\definecolor{MyBlue}{rgb}{0,0.2,0.8}
\definecolor{MyRed}{rgb}{0.8,0.2,0}
\definecolor{MyGreen}{rgb}{0.0,0.5,0.1}
\definecolor{MyGray}{rgb}{0.4,0.4,0.4}
\def\newtext#1{\textcolor{blue}{#1}}
\def\torevise#1{\textcolor{red}{#1}}
\def\red#1{\textcolor{red}{#1}}
\def\blue#1{\textcolor{blue}{#1}}
\def\green#1{\textcolor{green}{#1}}

\newcommand{\sE}{\mathbb{E}}

\newcommand\blfootnote[1]{%
  \begingroup
  \renewcommand\thefootnote{}\footnote{#1}%
  \addtocounter{footnote}{-1}%
  \endgroup
}

\newcommand{\cmark}{\ding{51}}%
\newcommand{\xmark}{\ding{55}}%

%% file: 1_Introduction.tex
\section{Introduction}
\label{sec:intro}

Given an incomplete image or a partial image input, humans generally are able to picture the context of the corresponding complete version. Such reasoning skill is largely based on our prior experience and knowledge observed from diverse images and their semantics. In the scope of machine learning, the objective is typically applied for the task of image completion, aiming to generate or predict reasonable missing image regions based on the observed input. In the areas of computer vision and image processing, several content creation applications such as object removal editing~\cite{suvorov2021resolution}, image panorama creation ~\cite{ying2020180}, texture creation~\cite{ulyanov2016texture}, and view expansion~\cite{yang2019very} are closely related to the aforementioned task. 

\begin{figure}[t]
\centering
    \includegraphics[width=0.8\linewidth]{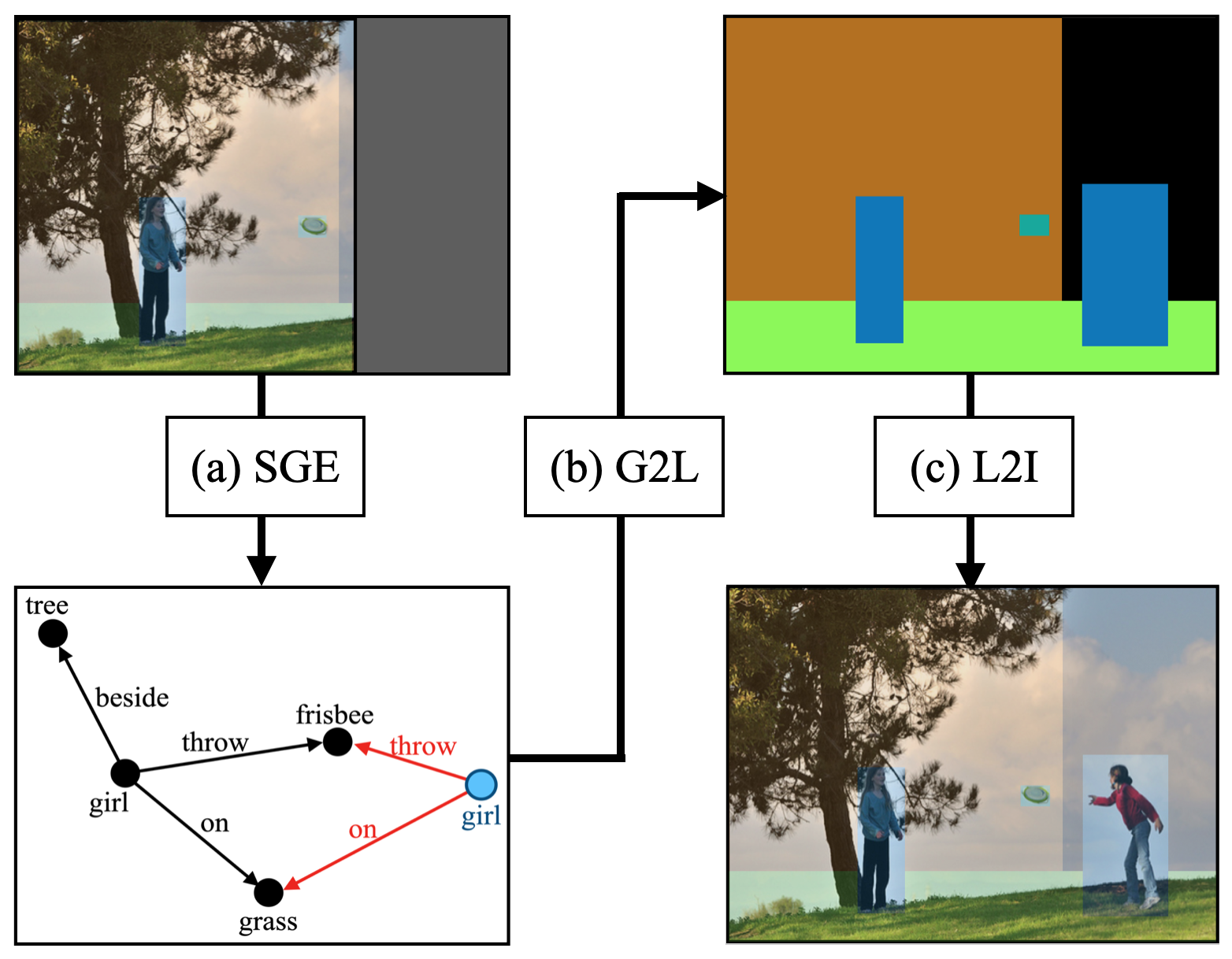}
    \vspace{-2mm}
\caption{\textbf{Illustration of semantics-guided image outpainting.} Our work can be divided into stages of (a) Scene Graph Expansion (SGE), (b) Scene Graph to Layout (G2L), and (c) Layout to Image (L2I) conversions. The blue node and red edges in the scene graph indicate the generated objects and relationship, respectively.}
\label{figure:teaser}
\vspace{-4mm}
\end{figure}

Depending on where the missing parts are to be recovered, the task of image completion is typically divided into two categories, image inpainting (also known as image hole-filling) and outpainting (also known as image extrapolation). Compared to image inpainting, image outpainting needs to synthesize unknown regions in single-sided fashions and thus is considered to be more challenging. Based on image inpainting works~\cite{iizuka2017globally, liu2018image, yu2019free, nazeri2019edgeconnect, xiong2019foreground}, researchers advance local and global GAN~\cite{iizuka2017globally}, Partial Convolution~\cite{liu2018image}, Gated Convolution~\cite{yu2019free} and edge information~\cite{nazeri2019edgeconnect} for outpainting tasks~\cite{lu2021bridging, sabini2018painting, yang2019very, wang2019wide, teterwak2019boundless, khurana2021semie}. However, despite impressive performances, most existing approaches are not designed to predict novel semantic regions in the output images. That is, they mainly focus on extending the surrounding texture or completing the fractional objects, resulting in extrapolated image regions with repeating structures or patterns. It is not clear how to introduce novel semantics with reasonable relationships with the existing ones during outpainting. As a result, we choose to approach this challenging semantics-oriented image outpainting problem by modeling and manipulating images at the semantic level. 

In order to tackle the above task, a scene graph would be a desirable representation due to their ability in describing the presence of semantic objects and their relationships in an image. Thus, based on recent works such as \cite{johnson2018image}, \cite{herzig2020learning} and \cite{park2019semantic}, one can describe and categorize a given image into three levels. The first level is the image level, containing pixel-level information. The second one is the layout level, which describes the locations/sizes of the objects of interest, including their corresponding category labels. The final level is the scene graph level, which describes semantic objects and their relationships (e.g., \textit{right of}, \textit{throw}) in an image. The higher the level is, the more abstract and semantic information it would contain. 

In this paper, we choose to decompose the semantics-guided image outpainting task into three stages, as depicted in \Figref{figure:teaser}. Given the scene graph extracted from the partial image and its layout, the first stage of scene graph expansion (SGE) utilizes the proposed Scene Graph Transformer (SGT), which uniquely performs node and edge-level attention, for expanding the input scene graph. The following stage of G2L further transforms such an expanded scene graph into a complete layout. Finally, layout-to-image (L2I) models can be applied for producing the final image output. We note that both SGE and G2L stages utilize our proposed SGT module, taking scene graph data as inputs with unique objectives introduced to enforce the desirable object/relationship properties, as later discussed in Sect.~\ref{sec:approach}.

The contributions of our work are highlighted as follows:
\begin{itemize}
\item We approach the task of semantics-guided image outpainting, which is able to synthesize novel yet semantically practical objects with associated relationships for completing an image output.

\item  We propose a Scene Graph Transformer (SGT), which takes node and edge features with unique node-level and edge-level attention mechanisms for modeling the associated structural information.

\item Expecting the sparsity of the object relationships in a scene graph, our SGT is designed to exploit the converse relationships between objects, so that semantically practical nodes and their corresponding edges can be properly recovered or expanded.

\end{itemize}

%% file: 2_related.tex
\section{Related Work}
\label{sec:related}

\subsection{Image Outpainting}

Adversarial learning~\cite{sabini2018painting} has been applied for image outpainting, generating image regions toward horizontal directions. By adopting a recurrent neural network, \cite{yang2019very} extends the output image in a single direction with varying lengths. As for \cite{lu2021bridging}, it fills in the intermediate gap between left and right partial image inputs for outpainting purposes. Although the method of \cite{wang2019wide} allows outpainting in all four directions, they require extra information (i.e., the image margins) during both training and testings. While such requirements are later alleviated by \cite{9506634}, most existing works are only capable of extending background textural regions or mending fractional objects. It is not clear whether novel yet semantically practical image regions can be added to the output image. Recently, \cite{khurana2021semie} proposes to outpaint images based on the extrapolated segmentation map, serving as guidance for generating novel object instances.

\subsection{From Scene Graphs to Images}
\label{sec:G2I}
As noted in \cite{johnson2015image}, scene graph is a data structure with each node encoding an object in the image, and each edge describing the associated relationship. Scene graph generation can be viewed as a task of image-to-text conversion. However, generating an image from a scene graph is a more challenging task, and is first tackled by \cite{johnson2018image} in an end-to-end learning fashion. Taking the image layout as an intermediate representation, one typically converts a scene graph to an image layout, followed by a layout-to-image conversion task. For scene graph to layout, \cite{herzig2020learning} leverages the converse and transition property of relationships. \cite{park2019semantic} proposes Spade, an architecture for describing image semantic layouts. \cite{herzig2020learning} extends Spade for manipulating the attributes of the generated objects. 

With the recent advances of the Transformer \cite{vaswani2017attention}, recent approaches like~\cite{yang2021layouttransformer, cong2021spatial, zareian2020learning} utilize Transformer based architectures for handling scene graph data, either for scene graph generation or scene graph to layout generation. However, these methods cannot be easily applied for scene graph expansion, which is critical in our focus on semantics-guided image outpainting. Nevertheless, since the Transformer deals with sequential data, one needs to convert the input scene graph to a sequence of triplets, each consisting of a \textit{subject}, a \textit{predicate}, and an \textit{object}. Moreover, since ``\textit{no relation}" would be also viewed as a predicate, describing a scene graph from an image would inevitably result in a large number of triplets. This would result in long triplet sequences, making the learning of the Transformer inefficient. Another potential problem for triplet representations is that if one object node has multiple relationship edges, the object node will appear in multiple triplets, which might result in redundant representations with inconsistent semantic outputs. 
In this paper, we propose an alternative yet novel architecture, Scene Graph Transformer (SGT). As detailed in the following section, our SGT would alleviate the aforementioned problems and can be applied for both scene graph expansion and scene graph to layout generation.

%% file: 3_approach.tex
\section{Methodology}
\label{sec:approach}

\begin{figure}[t!]
\centering
    \includegraphics[width=1.0\linewidth]{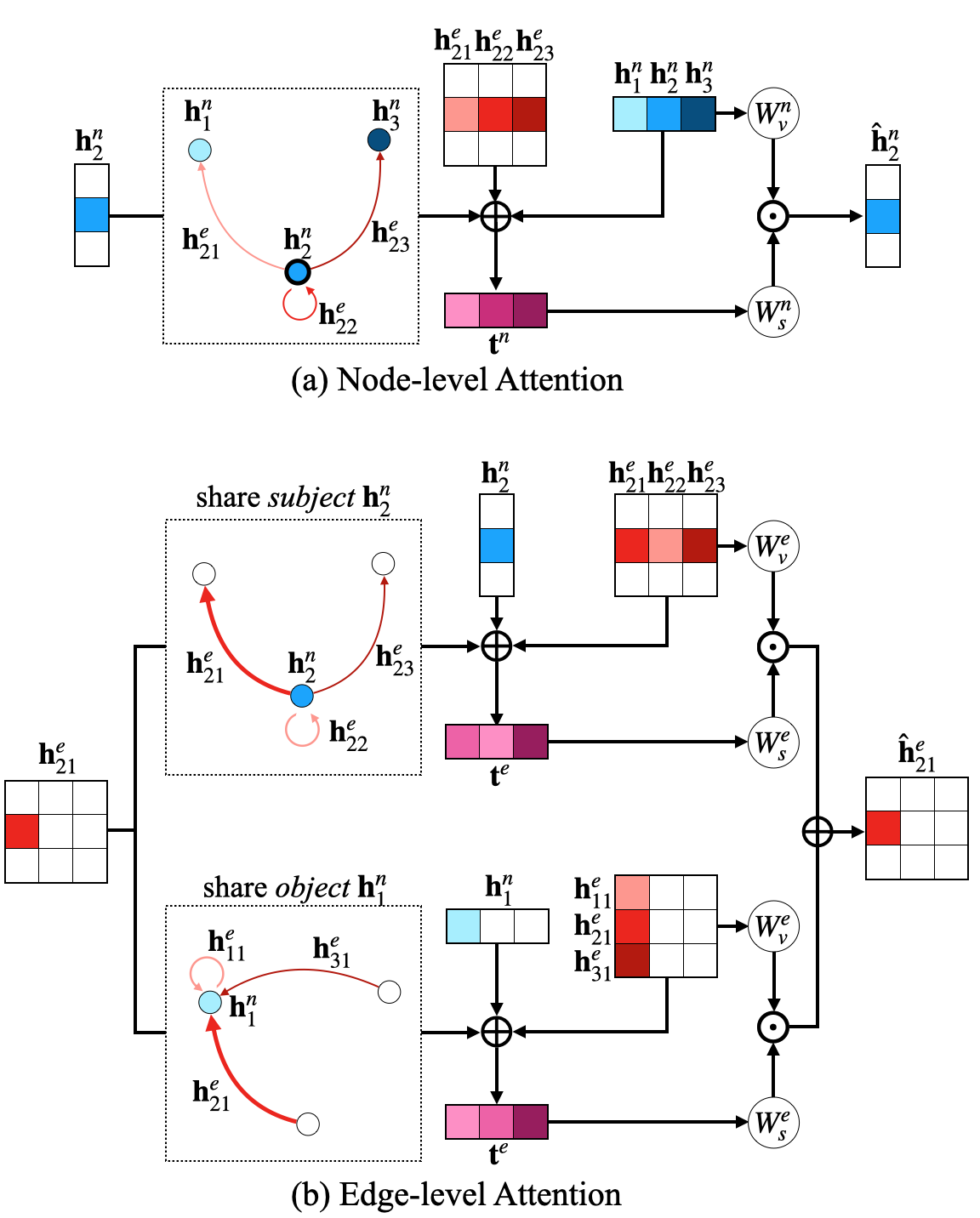}
\caption{\textbf{Scene Graph Transformer.} (a) Node-level attention: attention across nodes under guidance of the associated edges. (b) Edge-level attention: attention across edges conditioned on the sharing nodes. Note that $W^n_v, W^n_s, W^e_v$ $W^e_s$ denote the MLP transformation layers for the corresponding feature modalities. } 
\label{figure:scene_graph_transformer}
\vspace{-4mm}
\end{figure}

\subsection{Notations and Algorithmic Overview}
\label{sec:overview}
\textbf{Image outpainting.} Given an incomplete image $\mI^{in}$ of $h_1 \times w_1$ pixels, image outpainting is to generate an extrapolated image $\mI^{op}$ of $h_2 \times w_2$ pixels with $h_2 > h_1$ and $w_2 > w_1$. During training, we have $\mI^{in}$ partially cropped out from a complete image $\mI^{gt}$ (of $h_2 \times w_2$ pixels), aiming at producing $\mI^{op}$ to recover $\mI^{gt}$.

\textbf{Scene graph and layout.}
To describe semantic information in an image, a scene graph $\gS= (\mO, \mR)$ consists of a list of $N$ objects (nodes) $\mO = \{ o_i\}_{i=1:N}$ and the associated relationship (edge) matrix $\mR = (r_{ij}) \in \sR^{N \times N}$, where $o_i$ is the object label, and $r_{ij}$ indicates the edge label between objects $o_i$ and $o_j$. Note that $r_{ij}$ belongs to $\{ y^R_1, y^R_2, \cdots, y^R_{M} \} \cup \{0\}$, where each $y^R_i$ denotes the relation label (e.g., \textit{riding}, \textit{wear}, \textit{on}, etc.), and $M$ is its label number. And, $r_{ij} = 0$ indicates no relationship between the corresponding object pair. On the other hand, a layout is a list of bounding boxes of each object in an image, i.e., $\mathbf{B} = \{\vb_i\}_{i=1:N}$ with each $\vb_i = (b^x_i, b^y_i, b^w_i, b^h_i)$ describing the center coordinates and the size of the bounding box. We also compute the bounding box disparity for each relationship $\mathbf D = (\vd_{ij}) \in \sR^{N \times N \times 4}$, where each $\vd_{ij} = \{b_i^x - b_j^x, b_i^y - b_j^y, \log(b^w_i / b^w_j), \log(b^h_i) / \log(b^h_j)\}$ describes the spatial displacement between the bounding boxes of each subject-object pair.

\textbf{Algorithmic overview.} To perform semantic-guided image outpainting, our model would introduce novel object instances with realistic relationships with semantic practicality, which can be decomposed into the following three stages: 
The \textbf{scene graph expansion (SGE)} stage deploys the Scene Graph Transformer based on incomplete images $\mI^{in}$ with their layouts $\mL^{in}=(\mB^{in}, \mD^{in})$ and scene graphs $\gS^{in}=(\mO^{in}, \mR^{in})$, so that the model $T_{SGE}$ would expand $\gS^{in}$ into $\gS^{op}=(\mO^{op}, \mR^{op})$. In the stage of \textbf{scene graph to layout (G2L)}, we learn a second SGT-based model $T_{G2L}$ which converts the expanded scene graph into layout $L^{op}$ under the guidance of  $\mI^{in}$. Finally, for the \textbf{layout to image (L2I)} stage, we produce the final outpainted image $\mI^{op}$ via the model $G_{L2I}$. While not being the main focus of this work, our model $G_{L2I}$ is based on SPADE~\cite{park2019semantic} resnet blocks and consists of an image encoder and a generator.

\subsection{Scene Graph Transformer}
In this paper, we propose a novel architecture of Scene Graph Transformer (SGT), which is particularly designed to handle graph-structured data. With the ability to describe the nodes and their relationships in an image scene graph, our SGT performs separate yet mutually related self-attention between node levels and edge levels. That is, SGT views edges in scene graphs as regularization during the self-attention between different nodes, while the co-occurrence of nodes would guide the self-attention across different edges. Since both stages of SGE and G2L in our outpainting task take scene graph data as the inputs, our SGT will be utilized in both stages with objectives properly introduced and enforced.

For the sake of completeness, we briefly review the standard Transformer and explain how it can be applied to handle graph-structured data with $N$ nodes. As a sequence-to-sequence model, the Transformer consists of multiple transformation layers mapping an input sequence $\mH = \{\vh_i\}_{i=1:3N^2}$ to the output $\hat \mH = \{\hat \vh_i\}_{i=1:3N^2}$. Note that, with $N$ nodes and $N^2 = N \times N$ edges in the input graph, the Transformer in~\cite{yang2021layouttransformer, cong2021spatial, zareian2020learning} needs to convert such input data into a sequence, whose length is at least $3N^2$ due to the triplet representation ``\textit{subject}-\textit{predicate}-\textit{object}". For each transformer layer, the input vector $\vh$ is first converted into the query vector $\vq$, the key vector $\vk$, and the value vector $\vv$ through an MLP layer.
The output vector $\hat \vh$ is computed as the weighting sum of the value vector $\vv_j$, i.e., $\hat \vh_i = \sum_j s_{ij} \vv_j$, with the weight $s_{ij} = softmax(\vq_i \cdot  \vk_j / \sqrt{d_\mathbf{k}})$, where $d_\mathbf{k}$ is the dimension of $\vk$, and $\cdot$ stands for the inner product operation.

Instead of viewing the scene graph as a single sequence of triplets, the transformation layers in our SGT consider node (object) and edge (relationship) features as distinct yet mutually related data modalities. Thus, we have input and output of node feature sequences denoted as $\mH^n = \{\vh^n_{i}\}_{i=1:N}$ and $\hat \mH^n = \{\hat \vh^n_{i}\}_{i=1:N}$, respectively. As for those of the edge feature matrices, they are denoted as $\mH^e = \{\vh^e_{ij}\}_{i, j=1:N}$ and $\hat \mH^e = \{\hat \vh^e_{ij}\}_{i, j=1:N}$. For each modality, we deploy unique attention mechanisms based on the scene graph structure, as we present below.

\subsubsection{Node-level attention}
\label{section:node-attention}
The first type of attention in our SGT is performed at the node level, while the cross-attention between nodes is enforced by the edge relationships observed. Recall that, for the standard Transformer, it simply ``flattens'' the scene graph as a sequence of  (node$_i$-edge$_{ij}$-node$_j$) triplets with its attention mechanism not distinguishing between data modalities, nor considering the intrinsic graph structure. 

With the scene graph nodes as the inputs, our SGT calculates the similarity between node features $\vh^n_i$ and $\vh^n_j$ under the guidance of the associated edge feature $\vh^e_{ij}$, with the output for that node $\hat \vh^n_i$ as the weighted summation of the \textit{values} across each node $j$. 
Thus, we have
\begin{equation}
\label{equation:node-attention}
    \hat \vh^n_i = \sum_j \vs^n_{ij} \odot \vv^n_j,
\end{equation}
where $\vv^n_j$ is the value features for node $j$, $\vs^n_{ij}$ indicates the \textit{attention weight} derived from each triplet with node $i$ (i.e., node$_i$, edge$_{ij}$, and node$_j$), and $\odot$ denotes an element-wise multiplication.

As depicted in \Figref{figure:scene_graph_transformer}(a), the above calculation allows the edges associated with the node of interest to be incorporated into the attention process, which effectively regularizes the attention across nodes based on their corresponding relationships. To provide additional details, we calculate the value vector $\vv^n_i$ for each node $\vh^n_i$ through a single MLP $W^n_v$. Instead of utilizing query or key vectors for calculating the attention weight $\vs^n_{ij}$, we take the triplet features of \textit{node$_i$}-\textit{edge$_{ij}$}-\textit{node}$_j$ by concatenating their representations as $\vt^n_{ij} = \vh^n_i \oplus \vh^n_j \oplus \vh^e_{ij}$. With another MLP $W^n_s$ taking $\vt^n_{ij}$ as the input, the output weight vector $\vs^n_{ij}$ thus attends across $\vh^n_i$ and $\vh^n_j$, i.e. $\vs^n_{ij} = W_s^n(\vt^n_{ij})$. 

As a final remark, we do not follow the standard Transformer for using inner-product followed by \textit{softmax} to produce the attention weight. This is because our edge-regularized attention mechanism provides guidance of structural information, and the use of inner-product operation would dilute such information. Thus, we have the output vector $\hat \vh^n_i$ as the summation of element-wise multiplication between $\vs^n_{ij}$ and $\vv^n_j$, as shown in Equation~(\ref{equation:node-attention}).

\subsubsection{Edge-level attention}
\label{section:edge-attention}
For the edge features $\mH^e$ in a scene graph, only $\vh^e_{ij}$ contributes to the computation of $\vt^n_{ij} = \vh^n_i \oplus \vh^n_j \oplus \vh^e_{ij}$. If one simply performs cross-attention on $\vt^n_{ij}$, similar nodes would imply and result in the same $\hat \vh^e_{ij}$, which is viewed as the \textit{edge collapse} problem, i.e., resulting in repeating or redundant edges related to the same node $i$. For example, it is possible that, for node $j$ and node $k$ both linking to node $i$, the same $\hat \vh^e_{ij}$ and $\hat \vh^e_{ik}$ are produced (e.g., two \textit{men} hold the same \textit{tennis racket}).

To tackle the above problem, we propose edge-level attention in our SGT, while the cross-attention between edges is regularized by the nodes sharing the edge of interest, as illustrated in \Figref{figure:scene_graph_transformer}(b). To exploit inter-edge information and to take the shared nodes into consideration, we have an input edge feature $\vh^e_{ij}$ with the node pair $i, j$, and we consider edges linking to either node $i$ or $j$ for attention. Thus, we have the features $\vh^e_{kl}$ for such edges expressed as  $\{\vh^e_{kl} | k = i \vee l = j\}$.
And, the triplet feature for edge-level attention is computed as follows:
\begin{equation}
\label{eq:t^e}
\vt^e_{ij,kl} = \vh^e_{ij} \oplus \vh^e_{kl} \oplus 
    \begin{cases}
    \vh^n_i \text{, if } k = i
    \\
    \vh^n_j \text{, if } l = j.
    \end{cases}
\end{equation}
Instead of having the resulting edge-level attention matrix as $N^2 \times N^2 = N^4$, only $N^2$ (edge number) $\times$ $2N$ ($N$ \textit{subjects} $+ N$ \textit{objects})$= 2N^3$ edge pairs need to be considered. This greatly reduces the computation load when comparing to the use of the standard Transformer to perform attention across all edges in a graph. 

The remaining attention mechanism follows that of node-level attention discussed earlier. 
As depicted in \Figref{figure:scene_graph_transformer}(b), the above calculation allows the nodes associated with the edge of interest to be incorporated into the attention process, which effectively regularizes attention across edges based on their shared nodes. 
To provide further details, we calculate the value vector $\vv^e_{kl}$ for each edge $\vh^e_{kl}$ through a single MLP $W^e_v$. The ``edge triplet feature" of \textit{edge$_{ij}$}-\textit{shared node}-\textit{edge}$_{kl}$ is obtained by concatenating their representations as shown by~\Eqref{eq:t^e}. With an MLP $W^e_s$ taking $\vt^e_{ij,kl}$ as the input, the output weight vector $\vs^e_{ij,kl}$ thus attends across $h^e_{ij}$ and $h^e_{kl}$, i.e. $\vs^e_{ij,kl} = W_s^e(\vt^e_{ij,kl})$. 


\begin{figure*}[t!]
\centering
    \includegraphics[width=0.8\linewidth]{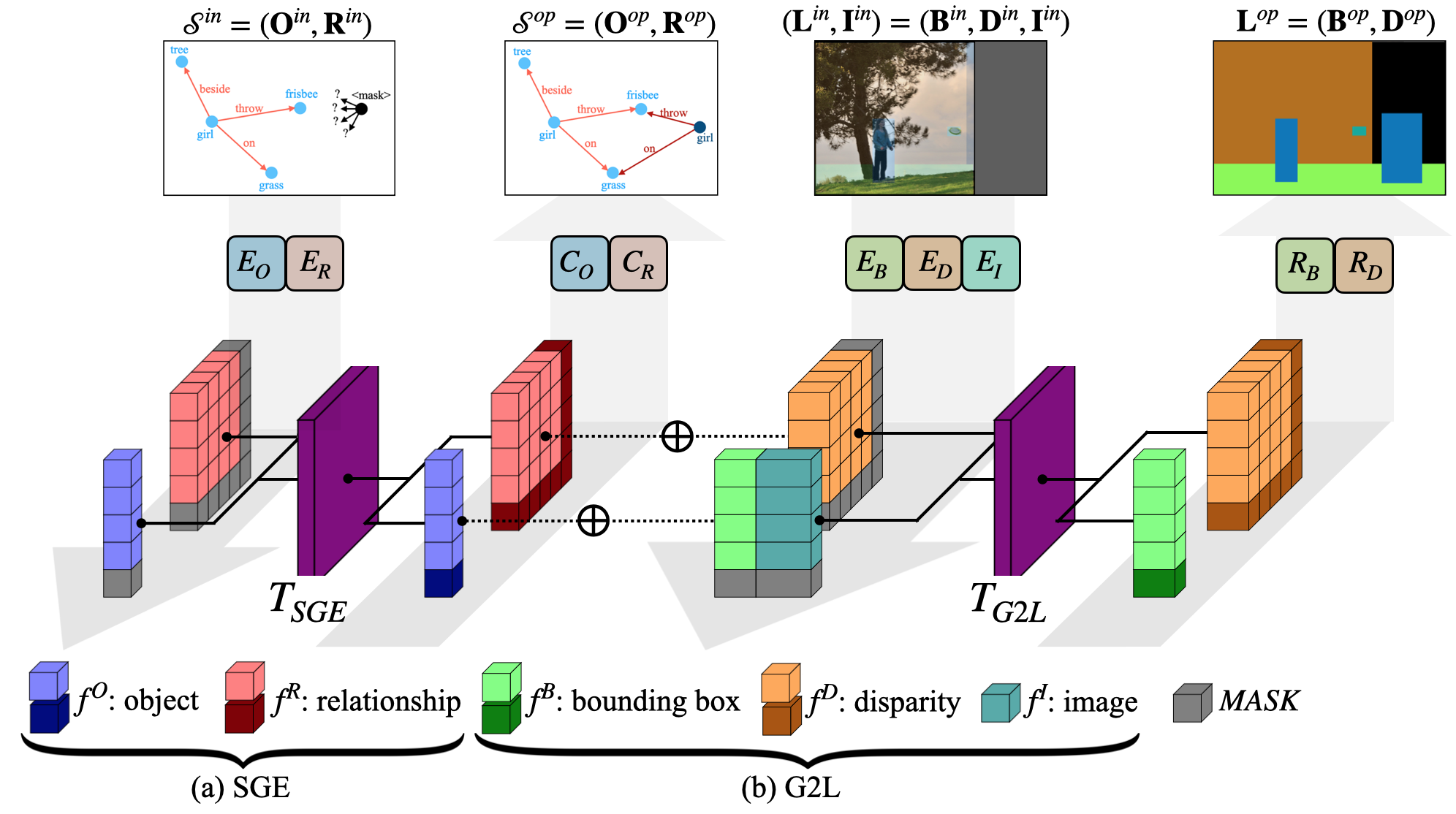}
    \vspace{-0.2cm}
\caption{\textbf{Flowcharts for (a) SGE and (b) G2L.} Note that the colors for the feature \textit{cubes} indicate their data modalities, and those in gray and dark-color denote the \textit{masked} and \textit{generated} ones. For SGE, the object and relationship classes are used as node and edge inputs, respectively. For G2L, we have the concatenated features of object class, bounding box, and the associated visual features as the node input, while the concatenated relationship label and its bounding box disparity as the edge inputs.} 
\label{figure:architecture}
\vspace{-4mm}
\end{figure*}

\subsection{Semantic-Guided Image Outpainting}

\subsubsection{Scene Graph Expansion}

Aiming at expanding the scene graph extracted from the input image, our SGT-based SGE model ${T}_{SGE}$ learns to append novel object nodes with the associated and necessary relationship edges introduced. Inspired by masked language model~\cite{devlin-etal-2019-bert}, we train this model by observing a complete scene graph $\gS^{gt} = (\mO^{gt}, \mR^{gt})$, with a number of objects in $\mO^{gt}$ being \textit{masked} with a special token [\textit{MASK}] assigned. Subsequently, the relationships in $\mR$ linking to the masked node (either a \textit{subject} or a \textit{object}) will also be \textit{masked}. This results in the partial input scene graph $\gS^{in} = (\mO^{in}, \mR^{in})$.

In order to perform node and edge-level attention, our SGE model ${T}_{SGE}$ contains object and relationship embedding encoders ${E}_O$ and ${E}_R$ to extract features from nodes and edges, object and relationship classifiers ${C}_O$ and ${C}_R$ for recognizing the derived output features, as shown in Fig.~\ref{figure:architecture}(a). That is, ${T}_{SGE}$ takes the object category word embeddings $\vf^O_i = {E}_O(o^{in}_i)$ as the node inputs $\vh^n_i$, and the relationship category word embedding $\vf^R_{ij} = {E}_R(r^{in}_{ij})$ as the relationship matrix input $\vh^e_{ij}$. 
${C}_O$ learns to predict the object class $o^{op}_i = C_O(\hat \vh^n_i)$, and the relationship classifier ${C}_R$ predicts the relationship label $r^{op}_{ij} = C_R(\hat \vh^e_{ij})$. From the above process, the objective of training ${T}_{SGE}$ is to recover the complete scene graph $\gS^{op} = (\mO^{op}, \mR^{op})$ from $\gS^{in}$. Thus, the objective can be summarized below:
\begin{equation}
\begin{split}
\label{eq:loss-SGE}
    \gL_{SGE} & = \sum_i \gL_{CE}(o^{op}_i, o^{gt}_i)  + \sum_{i,j} \gL_{CE}(r^{op}_{ij}, r^{gt}_{ij}),
\end{split}
\end{equation}
where $\gL_{CE}$ indicates the cross-entropy classification loss.\\


\textbf{Exploitation of converse relationships.} As noted in Sect.~\ref{sec:overview}, $\mR = (r_{ij}) \in \sR^{N \times N}$ denotes the relationships between each object pair in the scene graph. However, this matrix is \textit{not} necessarily expected to be a symmetric matrix, since $r_{ij} = y^R$ and $r_{ji} = \tilde{y}^R$ are viewed as relational antonyms and thus with \textit{converse relationships}, even both edges are connected to the same node pair. 
Given an input scene graph, typically only one of such relationship pairs would be observed. Thus, the above converse relationship can be implicitly inferred when either $r_{ij}$ or $r_{ji}$ is presented, resulting in the relationship matrix towards skew-symmetric (i.e., $r_{ji} = \tilde{r}_{ij}$). In practice, only a limited number of relationships would be specified in a scene graph, one thus observes a sparse ground truth relationship matrix $\mR^{gt}$, lacking converse relationship pairs. 
Also, one of the attention mechanisms introduced in our SGT is node-level attention, which is specifically guided by the relationship between the nodes of interest. Without properly generating and observing the aforementioned converse relationship pairs, the attention would be partially biased and result in undesirable outputs. The above challenges make the learning of the SGE model ${T}_{SGE}$ very difficult.

To tackle the above problem, we choose to process $\mR^{gt}$ as follows. For each non-\textit{empty} $r^{gt}_{ij}=y^R$, we manually assign the converse label $\tilde{y}^R$ to the associated empty $r^{gt}_{ji}$. (e.g., \textit{converse-riding} vs. \textit{riding}, and \textit{converse-on} vs. \textit{on}). It is worth noting that, the above label processing is for training ${T}_{SGE}$ only, not for later G2L and L2I training purposes.

Furthermore, to enforce the one-to-one mapping between a relationship and its converse version, we deploy an additional feature converter ${E}_C$ which takes the input relationship $E_R(y^R)$ and produces its converse version. This allows the classifier $C_R$ to predict its lalel $\tilde{y}^R$. Thus, $E_C$ is trained with the classification loss:
$\gL_{conv} = \sum_{i} \gL_{CE}({C}_R \circ {E}_C \circ {E}_R(y_i^R), \tilde{y}_i^R)$.

With converse relationship enforced, the skew-symmetry property of the SGE model can be expected, which can be calculated by the following loss function:
\begin{equation}
\vspace{-4mm}
\label{eq:loss-sym}
\gL_{sym} = \sum_{ij} \gL_{CE}({C}_R \circ {E}_C \circ {E}_R (r^{op}_{ij}), r_{ji}^{gt}).
\end{equation}

\noindent Note that $r^{op}_{ij} = C_R(\hat \vh^e_{ij})$ denotes the relationship label derived from the output relationship $\vh^e_{ij}$. Finally, our $T_{SGE}$ is trained with the combination of Equations~(\ref{eq:loss-SGE}) and~(\ref{eq:loss-sym}).

\begin{figure}[t]
\centering
\vspace{-4mm}
    \includegraphics[width=\linewidth]{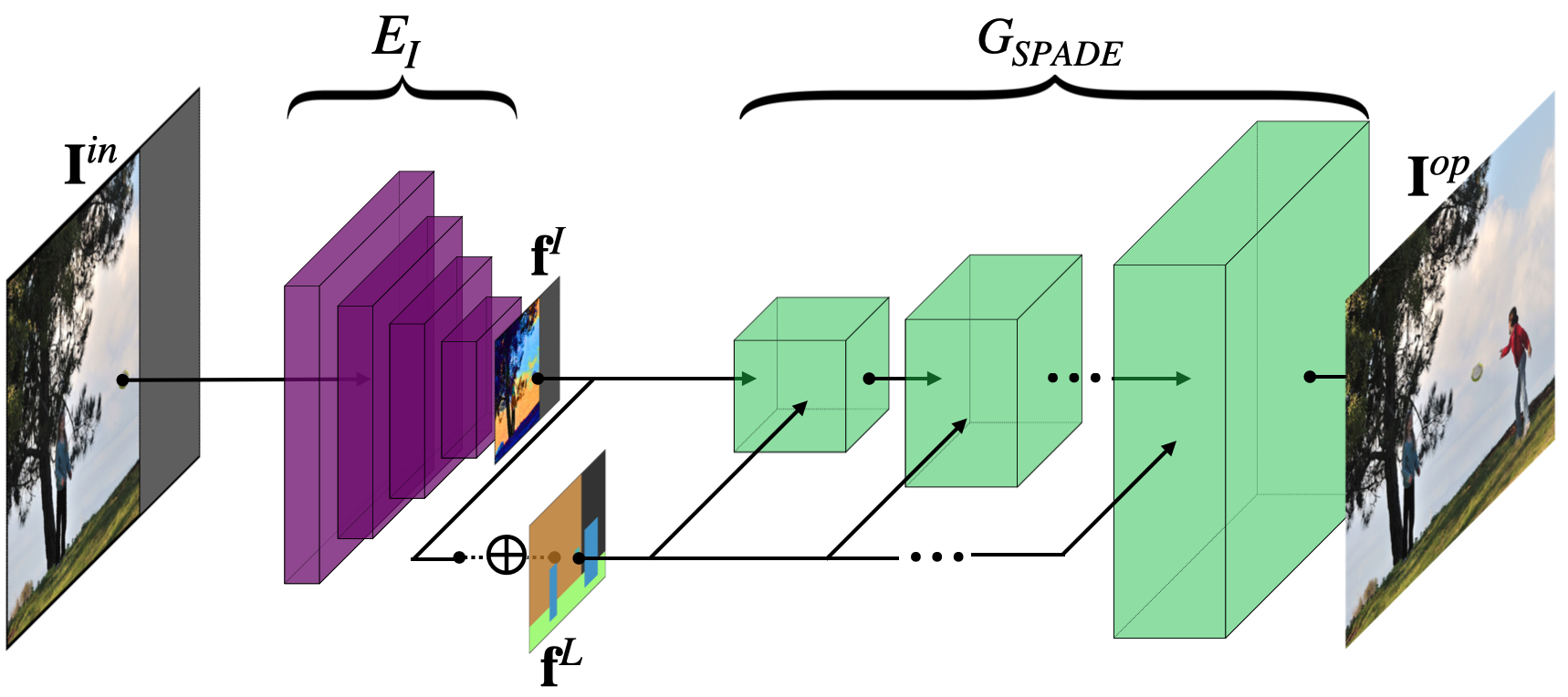}
\caption{\textbf{Illustration of our L2I Model $G_{L2I}$.} Based on AttSpade~\cite{herzig2020learning}, the decoder of $G_{L2I}$ (i.e., $G_{SPADE}$) takes semantic maps as guidance for producing image outputs.}
\label{figure:L2I}
\vspace{-6mm}
\end{figure}

\subsubsection{Scene Graph to Layout (G2L)}
\label{sec:G2L}
Given a partial input image $\mI^{in}$ with the corresponding layout $\mL^{in}$, together with the expanded scene graph $\gS^{op}$, the second stage of our work is to learn a G2L model $T_{G2L}$ for generating the plausible layout $\mL^{op}$. Based on the architecture of SGT but different from $T_{SGE}$, our $T_{G2L}$ considers a bounding box encoder $E_B$ with a regressor $R_B$, and a disparity encoder $E_D$ with a regressor $R_D$. Moreover, as shown in \Figref{figure:architecture}(b), an image encoder $E_I$ is deployed to distinguish whether a non-\textit{masked} object is with missing parts. For example, it is possible the $\mI_{in}$ consists of a horse with its legs cropped out of the image, thus with a smaller incomplete $\mL_{in}$. By feeding $T_{G2L}$ with the visual feature of $\mI_{in}$, it is expected to attend the partial horse and thus expand its incomplete bounding boxes accordingly.

Utilizing SGT, the input layout for $T_{G2L}$ is also described as a graph $(\mH^n, \mH^e)$. Each node $\vh^n_i$ is obtained by concatenating the object category embedding $\vf^{O}_i = E_O(o_i)$, bounding box feature $\vf^{B}_i = E_B(\vb_i)$, and the visual feature $\vf^{I}_i$, i.e. $\vh^n_i = \vf^O_i \oplus \vf^B_i \oplus \vf^I_i$. Note that $\vf^{I}_i$ can be directly obtained by cropping out the associated region from the input feature map $\vf^I = E_I(\mI)$. 
As for the edges in $\mH^e$, each edge input $\vh^e_{ij}$ is obtained by concatenating the relationship category embedding $\vf^{R}_{ij} = E_R(r_{ij})$ and the disparity feature $\vf^D_{ij} = E_D(\vd_{ij})$, i.e. $\vh^e_{ij} = \vf^{R}_{ij} \oplus \vf^D_{ij}$.

We note that the regressor $R_B$ in $T_{G2L}$ predicts bounding box information. If node $i$ denotes a novel/masked object, the regressor is trained to predict the bounding boxes $\vb^{op}_i$, under the supervision of ground truth $\vb^{gt}_i$. Otherwise, it would predict the boundary offset $\vb^{off}_i=\vb^{gt}_i - \vb^{in}_i$ (i.e., top, bottom, left, and right) between the input object and the ground truth one. As for the regressor $R_D$, it is deployed to maintain the consistency between the node outputs and edge outputs, and is trained to predict the bounding box disparities $\vd^{op}_{ij}$ under the supervision of ground truth $\vd^{gt}_{ij}$. With the above definitions, we train $T_{G2L}$ with the following loss:

\begin{equation}
\begin{split}
    & \gL_{G2L} = \sum_{i \text{, if } o^{in}_i \neq \textit{mask}} \gL_{cIoU}( \vb^{off}_i + \vb^{in}_i, \vb^{gt}_i  )
    \\ & + \sum_{i \text{, if } o^{in}_i = \textit{mask}} \gL_{cIoU} (\vb^{op}_i, \vb^{gt}_i) + \sum_{i,j} | \vd^{op}_{ij} - \vd^{gt}_{ij} |,
\end{split}
\end{equation}
where $\gL_{cIoU}(\cdot)$ is the complete-IoU loss utilized in~\cite{zheng2019distanceiou}.

\subsubsection{Layout to Image (L2I)}
With the expanded scene graph and layout, our final stage is to perform layout to image conversion. Adapted from AttSpade~\cite{herzig2020learning}, our L2I model $G_{L2I}$ learns to outpaint the partial input image into $\mI^{op}$, conditioned on $\gS^{op}$ and $\mL^{op}$. To enforce visual consistency, we choose to concatenate the image feature $\vf^{I} = {E}_{I}(\mI^{in})$ with the layout feature map $\vf^{L}$ to form the semantic information map $\vf^{S}$. This allows our model to generate a realistic output image with the guidance of $\vf^{S}$ through layers of SPADE blocks. Since the ground truth images are available during training, in addition to the adversarial loss, we are able to train $G^{L2I}$ with the reconstruction loss between $\mI^{op}$ and $\mI^{gt}$. 

It is worth repeating that, since we focus on the design of SGT (and its use for SGE and G2L), producing high-quality image outputs is \textit{not} within the main scope of this work. Thus, AttSpade-based designs can be replaced with state-of-the-art image conversion models if desirable.

%% file: 4_experiment.tex
\section{Experiments}
\label{sec:experiment}

\input{Tables/table_SGE}

\subsection{Datasets}

We evaluate our proposed methods on scene-level image datasets with bounding box annotation, namely COCO-stuff~\cite{lin2014microsoft, caesar2018coco}, VG-MSDN~\cite{krishna2017visual, li2017scene} and CityScapes~\cite{khurana2021semie}. Please see the supplementary materials for more details.

\blfootnote{The authors from NTU downloaded, evaluated, and completed the experiments on the datasets.}

\subsection{Evaluation and Analysis}
\label{sec:evaluation}

\textbf{Scene graph expansion.}
To compare the output expanded scene graph $\sS^{op}$ to the ground truth $\sS^{gt}$, we report the metrics of the averaged rank of correct prediction (rAVG) and the top-$k$ accuracy (Hits@$k$) for both object and relationship predictions, respectively. 
Note that We ignore the ``empty" relationship in $\gS^{gt}$ for accuracy calculation due to the sparsity expected for scene graphs. 

To assess that, compared to the training of masked language models (MLM), whether our proposed SGT learning strategy would favor the task of SGE, we consider/compare the following two training schemes. First (and as our proposed one), one object randomly is removed during training (including its corresponding edges). For the second case, we follow an existing MLM work~\cite{zareian2020learning} to randomly mask $\alpha = 30\%$ of the total objects and relationships for training the SGE model.

\begin{figure*}[t!]
\centering
\vspace{-8mm}
    \includegraphics[width=1\linewidth]{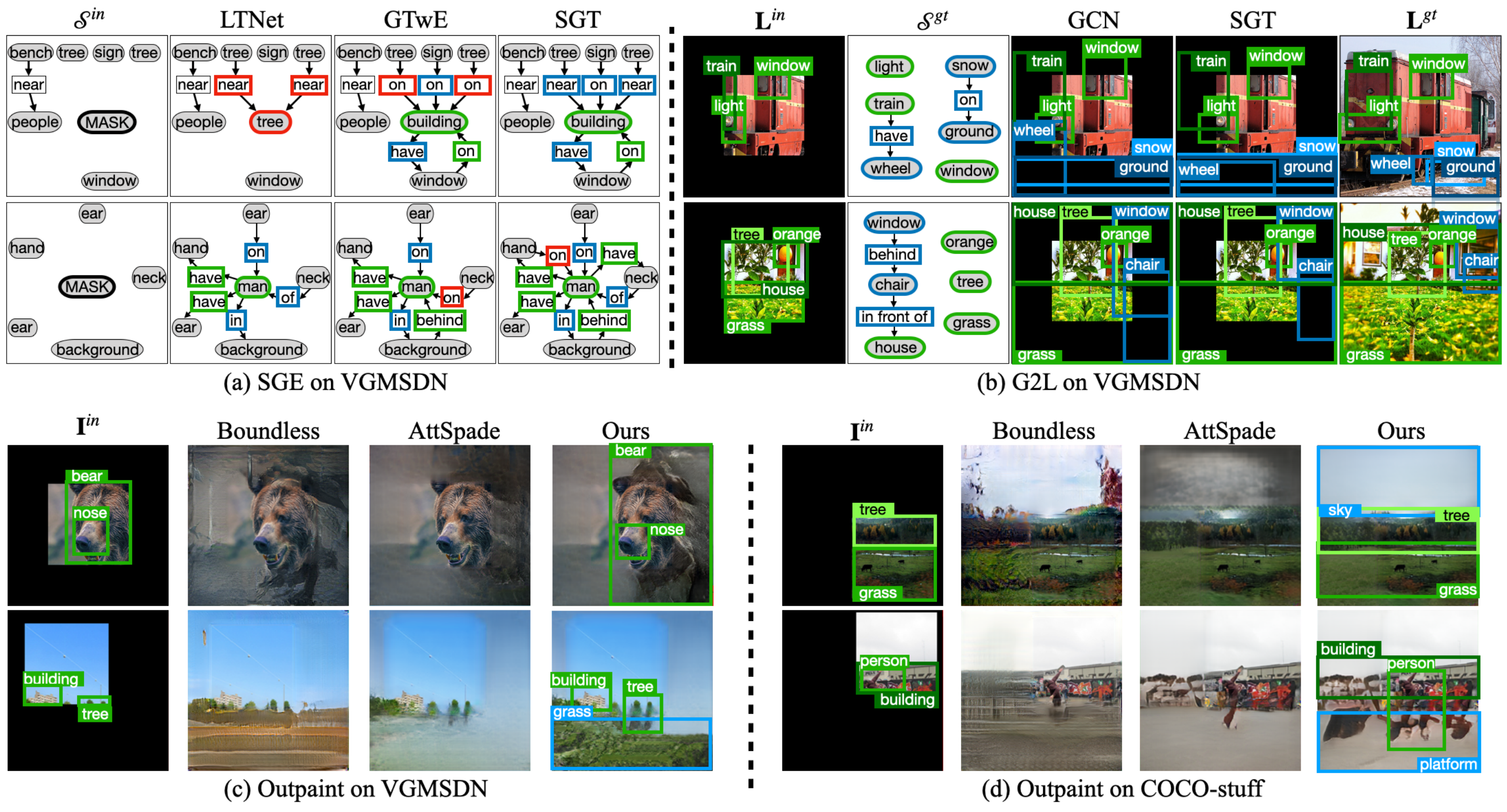}
\caption{\textbf{Visualization examples of SGE, G2L, and image outpainting.} From left to right: (a) input scene graph $\gS^{in}$, output scene graphs $\gS^{op}$ from LTNet, GTwE, and ours. Nodes and edges in green denote correct predictions, while those in blue are semantically practical but differ from the ground truth ones. Finally, those in red denote incorrect predictions. (b) input layout $\mL^{in}$, input scene graph $\gS^{gt}$, output layouts $\mL^{op}$ from GCN and Ours, and ground truth $\mL^{gt}$. Bounding boxes from novel (generated) objects are denoted in blue, while the existing ones are shown in green. (c) and (d): input image $\mI^{in}$, output images $\mI^{op}$ from Boundless, AttSpade, and ours. Note that we also highlight selected nodes and their bounding boxes following the protocol in (b).}
\label{figure:SGE-SG2L}
\vspace{-4mm}
\end{figure*}

In addition to the standard transformer, we apply LTNet~\cite{yang2021layouttransformer} and GTwE (Graph Transformer with Edge features~\cite{dwivedi2020generalization}) for comparison purposes, and the results are listed in Table~\ref{table:SGE}. From this table, it can be seen that our $T_{SGE}$ performed against baseline and state-of-the-art models on both VG-MSDN and COCO with clear margins, for both object generation and relationship prediction. More specifically, between the two training schemes, we found that the setting with a single object (and its edges) intentionally and randomly removed resulted in more effective performance. Since this setting is consistent with the expansion task for scene graphs, the use of SGT-based $T_{SGE}$ can be verified. Selected visualization examples of SGE are shown and compared in Fig.~\ref{figure:SGE-SG2L}(a).

\input{Tables/table_G2L}

\textbf{Scene graph to layout.} To evaluate the performance for this stage, we measure the mIoU between the output layout $\mL^{op}$ and the ground truth $\mL^{gt}$. We note that depending on the object is an introduced or an existing one, we show their mIoU separately, with the total mIoU as their weighted average. We compare our SGT-based $T_{G2L}$ with two other models: Transformer~\cite{vaswani2017attention}, GCN~\cite{kipf2016semi, herzig2020learning} and GTwE~\cite{dwivedi2020generalization}. Table \ref{table:G2L} lists the performances of the above methods, and we see that our model consistently performed against GCN and Transformer across different settings. From the visual examples shown in Figure~\ref{figure:SGE-SG2L}, it can be observed that our model better comprehends the expanded scene graph and the given partial input, so that the predicted layout would be more semantically practical.

Furthermore, we demonstrate the robustness of our model which is trained without any input image or layout guidance. This is thus consistent with the setting used in Sg2Im~\cite{johnson2018image}, Canonical~\cite{herzig2020learning}, and LTNet~\cite{yang2021layouttransformer}. With no input guidance except for our expanded scene graphs, all objects would be considered \textit{masked}(new) and thus only one mIoU score is reported. From the results shown in Table \ref{table:G2L}, one can see that our model still achieved the highest mIoU and thus would be preferable in such scenarios.

\textbf{Semantic-guided image outpainting.}
Finally, we evaluate the performance for semantics-guided image outpainting. From the visual examples shown in~\Figref{figure:SGE-SG2L}(c) and (d). We see that, under the guidance of the expanded scene graph and completed layout, our model better generates novel object instances at the pixel level, e.g., a \textit{grassland} (second row of Fig.~\ref{figure:SGE-SG2L}(c)), \textit{sky} (first row of Fig.~\ref{figure:SGE-SG2L}(d)), or expands an existing object to a reasonable size, e.g., the jaw of the bear (first row of Fig.~\ref{figure:SGE-SG2L}(c)). Additional visualization results are provided in the supplementary materials.

We note that, most image outpainting works consider images in restrained scenes (e.g., Cityscapes~\cite{khurana2021semie} and ADE20K~\cite{khurana2021semie}) or those with single category object (e.g., CUB~\cite{wang2019wide}, CelebA~\cite{wang2019wide, zhang2020nested} or DeepFashion~\cite{wang2019wide}). And, to the best of our knowledge, we are the first outpaint image data in the wild with rich interaction between various categories of objects (e.g., VG-MSDN and COCO-stuff). Therefore, only limited quantitative comparisons can be conducted. Specifically, we consider Cityscapes~\cite{Cordts2016Cityscapes} and take the Fréchet inception distance (FID)~\cite{heusel2017gans} as the metric. Our model reported FID of 60.99, which surpassed Outpainting-SRN~\cite{wang2019wide} at 66.89, Boundless~\cite{teterwak2019boundless} at 77.86, and a modified AttSpade~\cite{herzig2020learning} at 68.91 (equivalent to our $G_{L2I}$ only). While a recent work of SemIE~\cite{khurana2021semie} reported an improved FID score of 47.67, it is designed for restrained street-view (Cityscapes) or indoor scenes (ADE20K~\cite{zhou2017scene}), and cannot be easily applied to outpaint image data in the wild as ours does. Another requirement of SemIE is the use of segmentation masks as learning guidance, while we only require guidance at scene graph levels. Thus, the effectiveness and practicality of our proposed model can be verified.

\textbf{Ablation studies.} To assess the design of our SGT, we consider VG-MSDN and report the performance on SGE. For Hits@$1$, the baseline SGT with only node-level attention reported 35.7/46.1 on object/relationship prediction while adding edge-level attention and regularization of skew-symmetry result in 38.7/48.6 and 38.2/52.0, respectively. Finally, our SGT with full objectives achieved 39.7/55.3, which confirms its design and learning schemes. More details can be found in the supplementary materials.

%% file: Tables/table_SGE.tex
\begin{table*}[t]
\centering
\small
\caption{\textbf{Quantitative evaluation on scene graph expansion.} Note that masking strategies of M and E denote uses of standard MLM and our expansion-based learning schemes, as described in Sect. 4.2.}
\label{table:SGE}
\captionsetup{singlelinecheck = false}

\begin{tabular}{ l c c c c c  c c c c } 
\toprule
{} &  & \multicolumn{4}{c}{\textbf{VG-MSDN}} & \multicolumn{4}{c}{\textbf{COCO-stuff}} \\ 
\cmidrule(lr){3-6}
\cmidrule(lr){7-10}
 & \multirowcell{2}{Masking \\ strategy} & \multicolumn{2}{c}{Object} & \multicolumn{2}{c}{Relation} & \multicolumn{2}{c}{Object} & \multicolumn{2}{c}{Relation} \\
\cmidrule(lr){3-4}
\cmidrule(lr){5-6}
\cmidrule(lr){7-8}
\cmidrule(lr){9-10}
{} & & {rAVG $\downarrow$} & {Hit@ 1 / 5 $\uparrow$} & {rAVG $\downarrow$} & {Hit@ 1 / 5 $\uparrow$} & {rAVG $\downarrow$} & {Hit@ 1 / 5 $\uparrow$} & {rAVG $\downarrow$} & {Hit@ 1 / 3 $\uparrow$} \\

\midrule
{Transformer} & M & {28.96} & {9.55 / 29.7} & {5.32} & {37.4 / 70.1} & {31.14} & {11.1 / 29.7} & {2.41} & {28.7 / 78.2} \\
{LTNet} & M & {24.72} & {9.86 / 36.4} & {4.62} & {42.7 / 74.8} & {30.76} & {12.0 / 30.3} & {2.40} & {27.9 / 77.8}\\ 
{GTwE} & M & {10.93} & {28.3 / 58.8} & {5.26} & {34.9 / 73.1} & {12.25} & {24.6 / 54.1} & {2.92} & {20.1 / 61.9}  \\
{SGT} & M & {9.40} & {34.7 / 64.5} & {3.92} & {48.7 / 81.7} & {11.32} & {26.0 / 58.5} & {2.20} & {36.4 / 81.5}
 \\

\midrule
{Transformer} & E & {33.77} & {10.6 / 28.9} & {5.30}  & {35.3 / 65.8} &  {22.35} & {14.7 / 37.8} & 2.37 & {29.4 / 78.5}\\
{LTNet} & E & {24.45} & {13.9 / 34.8} & {4.70} & {34.8 / 74.6} & {17.22} & {20.1 / 45.8} & {2.36} & {29.1 /  78.4}\\ 
{GTwE} & E & {11.91} & {27.0 / 57.2} & {5.36} & {35.8 / 72.5} & {11.81} & {28.4 / 57.2} & {2.89} & {20.4 / 63.3}\\
{SGT} & E & \textbf{8.38} & \textbf{39.7 / 68.9} &  \textbf{3.43} & \textbf{55.3 / 84.3} & \textbf{11.03} & \textbf{29.6 / 59.0} & \textbf{2.19} & \textbf{45.5 / 82.2}\\

\bottomrule
\end{tabular}
\end{table*}

%% file: Tables/table_G2L.tex
\begin{table}[t]
\small
\centering
\caption{\textbf{Quantitative Results on scene graph to layout.} The (\cmark / \xmark) indicates whether the method is trained and test with or without input image and layout guidance.}
\label{table:G2L}
\begin{tabular}{l c c c} 
\toprule
{} &  & {\textbf{VG-MSDN}} & {\textbf{COCO-stuff}} \\ 
\cmidrule(lr){3-3}
\cmidrule(lr){4-4}
{} & {$\mL^{in}$} & {mIoU} & {mIoU} \\

\midrule
{Sg2Im} & \xmark & {16.8} & {29.0}  \\
{Canonical} & \xmark & {18.0} & {41.9} \\ 
{LTNet} & \xmark & {18.3} & {49.0}  \\
{SGT} & \xmark & {\textbf{25.3}} & {\textbf{54.6}} \\

\midrule
{Transformer} & \cmark & { 5.1 / 71.2 / 51.9} & {10.4 / 75.7 / 61.2} \\
{GCN} & \cmark & {11.4 / 70.6 / 50.0} & {21.1 / 72.3 / 60.8}\\
{GTwE} & \cmark & {12.3 / 79.9 / 62.1} & {21.3 / 73.2 / 64.8} \\
{SGT} & \cmark & \textbf{14.5 / 81.1 / 62.4} & \textbf{28.2 / 85.1 / 74.9} \\

\bottomrule
\end{tabular}
\vspace{-4mm}

\end{table}

%% file: 5_conclusion.tex
\section{Conclusions}
\label{sec:conclusion}
We address the task of semantics-guided image outpainting by proposing a novel Scene Graph Transformer (SGT). By decomposing the task into the stages SGE, G2L, and L2I, our proposed model leverages information observed from the nodes and edges in the partial input scene graph, inferring plausible object co-occurrences, and thus producing the final image output. Our SGT uniquely performs attention at both node and edge levels for modeling input structural information. In addition, for completing a semantically practical image, our SGT exploits converse relationships between edges for scene graph expansion. Our experiments confirmed that our proposed SGT performs favorably against state-of-the-art transformer-based models on both SGE and G2L. With novel objects and their relationships introduced, satisfactory image outputs can be achieved.

%% file: 7_acknowledgement.tex
\paragraph{Acknowledgement}

This work is supported in part by the Ministry of Science and Technology of Taiwan under grant MOST 110-2634-F-002-036 and  in part by Qualcomm Technologies, Inc. through a Taiwan University Research Collaboration Project. We also thank to National Center for High-performance Computing (NCHC) for providing computational and storage resources.

%% file: 6_supplementary.tex
\appendix
\noindent {\Large \textbf{Supplementary Material}}

\section{Implementation}

\subsection{Architecture}

\noindent \textbf{Encoders $E_O, E_R$ and Classifiers $C_O, C_R$ for SGE.} Both the object embedding encoder $E_O$ and the relationship embedding encoder $E_R$ are word embedding codebooks. The object classifier $C_O$ has its weight shared with $E_O$. That is, when classifying an output feature $\vf^O_i$, the classifier will compare it to each object codevector in the codebook. The object class with the highest cosine similarity between the codevector and the feature is then returned as predicted label.
The classifier $C_R$ has its weight shared with $E_R$ for the same reason.\\

\noindent \textbf{Encoders $E_B, E_D$ and Regressors $R_B, R_D$ for G2L.} Both the bounding box encoder $E_B$ and the disparity encoder $E_D$ are two-layer MLPs. The disparity regressor $R_D$ is a two-layer MLP while the bounding box regressor $R_B$ contains two distinct two-layer MLPs, $R_B^1$ and $R_B^2$ respectively. As mentioned in Sect. \red{3.3.2}, one of the MLP $R_B^1$ is trained for predicting the bounding boxes for novel/masked objects and the another one $R_B^2$ is trained for predicting the boundary offset for existing objects.\\

\noindent \textbf{Encoder $E_I$ for G2L \& L2I.} The architecture of the image encoder $E_I$ we deployed is detailed in Table~\ref{table:E_I}. The $E_I$ consists of five 2d convolution layers with batch normalization and leaky relu activation in each layer.\\ 

\begin{figure}[t!]
\centering
    \includegraphics[width=0.8\columnwidth]{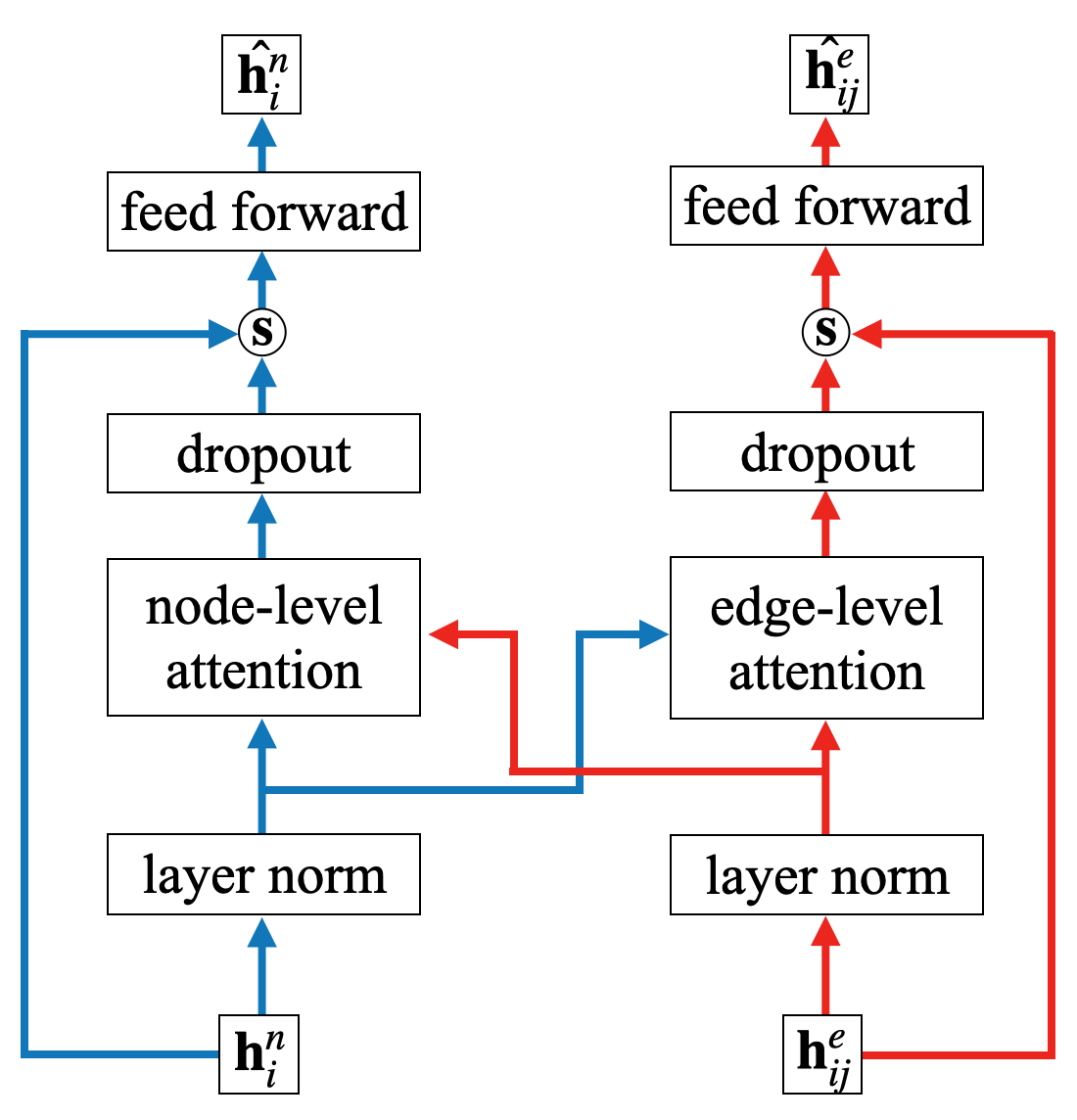}
\caption{\textbf{The block diagram for our proposed SGT layer.} The blue arrows indicate the flow of the node inputs $\vh^n$, while the red ones denote the flow of the edge inputs $\vh^e$. In order to exploit the observed graph structure, the node-level attention of $\vh^n$ in our SGT is guided by the edge inputs $\vh^e$, while the edge-level attention of $\vh^e$ is guided by the node inputs $\vh^n$. The symbol $\vs$ indicates the summation operation.}
\label{figure:SGT-detail}
\vspace{-6mm}
\end{figure}

\noindent \textbf{Scene Graph Transformers for SGE \& G2L.} The architecture overview of the SGT layer is shown in \Figref{figure:SGT-detail}. Noted that the details of our node-level and edge-level attention are depicted in Sect. \red{3.2.1} and Sect. \red{3.3.2}. 
All the SGT layer in $T_{SGE}$ and $T_{G2L}$ consists of attention hidden size $d_{atten}=512$, feed forward size $d_{ff}=2048$, multi-head number $n_{head}=4$, and dropout $=0.1$. Both $T_{SGE}$ and $T_{G2L}$ have 4 SGT layers.\\

\noindent \textbf{Generator $G_{SPADE}$ for L2I.} Our $G_{SPADE}$ is modified from AttSpade~\cite{herzig2020learning}, with the ability to take input image as guidance for image outpainting work. The generator is comprised of seven SPADE Resnet blocks. Apart from the size of input channels, which is required to be adjusted due to the input image feature, the architectures and the parameters mostly follow their original setting with the same Pix2Pix models as image and object discriminators. Please refer to the official repository \url{https://github.com/roeiherz/CanonicalSg2Im} for more details of the implementation of AttSpade.

\begin{table}[t]
\vspace{-2mm}
\caption{Architecture design of image encoder $E_I$ for the stages of G2L and L2I.}
\centering
\begin{tabular}{l l} 

\toprule
Type & Argument \\
\midrule
{Conv2d} & {c=64, k=5, pad=2} \\
{BatchNorm2d} & {c=64} \\
{LeackReLU} & {slope=0.2} \\

{Conv2d} & {c=128, k=3, stride=2, pad=1} \\
{BatchNorm2d} & {c=128} \\
{LeackReLU} & {slope=0.2} \\

{Conv2d} & {c=128, k=3, stride=2, pad=1} \\
{BatchNorm2d} & {c=128} \\
{LeackReLU} & {slope=0.2} \\

{Conv2d} & {c=256, k=3, stride=2, pad=1} \\
{BatchNorm2d} & {c=256} \\
{LeackReLU} & {slope=0.2} \\

{Conv2d} & {c=256, k=3, pad=1} \\
{BatchNorm2d} & {c=256} \\
{LeackReLU} & {slope=0.2} \\

\bottomrule
\end{tabular}
\vspace{-4mm}
\label{table:E_I}
\end{table}

\noindent \textbf{Baselines.}  For the experiments in Table~\red{1} and Table~\red{2}, the Transformer, LTNet and GTwE are all set to have the same aforementioned hyper-parameters, which are the attention hidden size $d_{atten}=512$, feed forward size $d_{ff}=2048$, multi-head number $n_{head}=4$, and dropout $=0.1$. For the experiments in Table~\red{2}, the hidden size of GCN is also set to $512$.

\subsection{Datasets}
\noindent \textbf{COCO-stuff.} The COCO dataset~\cite{lin2014microsoft} provides images of 80 object categories with bounding box information. COCO-stuff~\cite{caesar2018coco} adds 91 more \textit{stuff} categories (e.g., sky, snow, etc.), with 118K/50K images for training and validation.

\noindent \textbf{VG-MSDN.} 
The original Visual Genome dataset~\cite{krishna2017visual} contains more than 100K images with noisy labels, and thus we consider its subset of VG-MSDN~\cite{li2017scene}, containing 150 object categories and 50 relationship categories with about 45K/10K for training/validation.

\noindent \textbf{Cityscapes.} 
The Cityscapes dataset~\cite{khurana2021semie} contains 2,975/500 images for training and validation.
It provides about 40 object categories with bounding box information for each image.

\subsection{Scene Graph Generation}
For datasets with scene graph annotation, such as VG-MSDN, the labeled scene graphs are used as ground truth without any pruning. For datasets without scene graph annotation, such as COCO and CityScapes, we follow the processing technique of~\cite{herzig2020learning, yang2021layouttransformer} to generate ground truth scene graphs. That is, we utilize the ground truth objects \& bounding boxes to construct rule-based scene graphs. This processing stage does \textit{not} require extra data/label supervision.

\subsection{Training and Inference}
All our models are trained with Adam optimizer and the base learning rate $\gamma$ is set to 4e-4.
We train the three models ($T_{SGE}$, $T_{G2L}$, and $G_{L2I}$) individually rather than end-to-end, since the errors of upper-stream models can greatly deteriorate the result of down-stream ones during training, i.e. incorrect generated novel object in SGE leads to impossible image reconstruction in L2I.

\subsubsection{Scene Graph Expansion}
We train our SGE model for a total of 10 epochs. We deploy the learning rate schedule as the following. The initial learning rate $\gamma_0$ is set to a tenth of $\gamma$. In the warmup stage, the learning rate grows linearly up to the maximum learning rate $\gamma$. Then the learning rate will hold for a short period of steps. Finally, the learning rate decay exponentially until the minimum learning rate $\num{2.5e-5} \gamma$ is hit at the same time as the end of training. The warmup stage spans a tenth of our total training steps while the hold stage spans a thousandth. It takes 4-12 hours to train on a single NVIDIA-GTX 1080 depending on the datasets and different hyper-parameters.

For data processing on scene graphs, we also introduced a few dummy tokens besides [\textit{MASK}] (the token for \textit{masked} objects and relationships).
A dummy object [\textit{IMAGE}] is added to prevent null input.  Each object will have an relationship [\textit{IN\_IMAGE}] between it and the dummy object [\textit{IMAGE}]. Each object will have an relationship [\textit{SELF}] between it and itself, i.e. $\forall i \in 1:N, r^{gt}_{ii} = \textit{SELF}$. Noted that when considering the objective function for training, such target labels are not excluded from computation. 

We follow the setting in~\cite{yang2021layouttransformer} where images with less than 3 objects are excluded while images with more than 8 objects will have the exceeding objects ignored. That is, if there are 10 objects in the given scene graph, we will randomly keep 8 of all. This rule is also applied to the training and inference on G2L, while on L2I the maximum number of objects is set to 30.

The objective function $\gL_{SGE}$ has to be further adjusted since the distribution of the relationship labels is highly uneven, which can have a negative impact on training. That is, when computing the objective function $\gL_{SGE}^r = \gL_{CE}(r^{op}_{ij}, r^{gt}_{ij})$, most of the target $r^{gt}_{ij}$ is \textit{no-relation} due to the sparsity of the annotation in the datasets. To prevent biased training, weighted cross-entropy loss is deployed instead, and the weight of label \textit{no-relation} is set to $0.05$ while the weights of other labels are still set to $1.0$.

\subsubsection{Scene Graph to Layout}
Each ground truth image $\mI^{gt}$ is first resized to 256 $\times$ 256 pixels and a sub-image of size 128 $\times$ 128 pixels is cropped out as the input image $\mI^{in}$. The images then are normalized to the range $[-1, 1]$. The associated mask $\mM^{in}$ indicating which part of the $\mI^{in}$ is input guidance and which part is missing region will also be concatenated onto $\mI^{in}$ as the input to $E_I$.
Objects that have their bounding boxes completely out of the cropped region will be considered \textit{masked} and so are the associated relationships. 

As for the input bounding boxes $\vb^{in}$, those of the \textit{masked} objects will be set to $(b^x, b^y, b^w, b^h) = (0.5, 0.5, 0.0, 0.0)$. The bounding box of the dummy object [\textit{IMAGE}] is set to $(0.5, 0.5, 1.0, 1.0)$. The rest of the objects will have their bounding boxes reduced to be in the area of the cropped region. All the input disparities $\vd^{in}$ will be calculated after the processing on $\vb^{in}$.

We train our G2L model for a total of 50 epochs. The learning rate schedule is the same as the one used for SGE. It takes about 2 days to train on a single NVIDIA-GTX 1080 depending on the datasets and different hyper-parameters.

\subsubsection{Layout to Image}
The image pre-processing is the same as G2L.
We train our L2I model for a total of 30 epochs. No learning rate schedule is used. It takes about 7 days to train on a single NVIDIA-V100.

\section{Ablation Studies}

Due to the limitation of spaces, we report only a part of the scores for ablation studies in Sect.~\red{4.2}. Here we report the complete table with both rAVG and Hit@1/3 as metrics for ablation studies on SGE in Table.~\ref{table:ablation}. It is worth noting that, since scene graphs are often sparse, adding only edge-level attention w/o exploiting converse relationships, i.e. $L_{sym}$, tends to overfit the observed relationships thus with degraded results.
\input{Tables/table_ablation}

Additionally, we also conduct ablation studies for G2L to evaluate the effectiveness of our design. As shown in Table~\ref{table:ablation-G2L}, the introduction of edge-level attention is also beneficial to the layout prediction. Adding in visual features $\vf^I$ from $E_I$ greatly enhances the performance on layout expansion with existing objects. Overall, with full objective implemented, our model achieve the best performance.

\begin{table}
\caption{\textbf{Ablation studies of our SGT for scene graph to layout on VG-MSDN in terms of mIoU.} The three mIoU scores in each row are calculated for novel objects, existing objects, and all objects, respectively.}
\label{table:ablation-G2L}
\centering
\begin{tabular}{ l c } 
\toprule
{} & {mIoU} \\ 
\midrule
{node-level} & 11.5 / 79.2 / 61.1 \\
\midrule
{+ edge-level} & 12.0 / 80.0 / {62.1} \\
{+ $E_I$} & {{13.2} / {80.3} / 60.8} \\
\midrule
{Ours} & \textbf{14.5} / \textbf{81.1} / \textbf{62.4} \\
\bottomrule
\vspace{-4mm}
\end{tabular}
\end{table}

\section{Quantitative Result}
For image outpainting on VG-MSDN and COCO-stuff, we report the FID scores (the lower the better) for evaluation of the quality of outpainted images. As shown in Table~\ref{table:FID}, our model surpass Boundless~\cite{teterwak2019boundless} by a great margin on VG-MSDN. It is also expected that our model is only slightly better than AttSpade due to the similarity between our L2I architecture and theirs. The main difference lies on the fact that our model is able to generate novel objects which itself does not directly enhance image quality but resulting in richer and more meaningful images. Our advantage is reflected better in Sect.~\ref{sec:visual-comparison}

\begin{table}[t]
\centering
\caption{\textbf{Quantative result of image outpainting on VG-MSDN in terms of FID.}}
\label{table:FID}
\begin{tabular}{l c} 
\toprule
{} &  {\textbf{VG-MSDN}}\\ 
{} & {FID} \\

\midrule
Boundless~\cite{teterwak2019boundless} & 36.07  \\
AttSpade~\cite{herzig2020learning} & 23.61 \\
\midrule
{Ours} & \textbf{23.42} \\ 
\bottomrule
\end{tabular}
\vspace{-4mm}
\end{table}

\section{Visualization}

\subsection{t-SNE on relationship features}
To examine the effect of our proposed feature converter $E_C$ for exploiting converse relationships on VG-MSDN, we project all the relationship features $E_R(y^R_i)$ and converse relationship features $E_R(\tilde{y}^R_i)$ for all $i = 1:M$ onto a 2D plane with t-SNE, where each $y^R_i$ is a relationship label while $\tilde{y}^R_i$ is its pseudo converse label, and $M$ is the number of relationship labels.

\begin{figure}[t!]
\centering
    \includegraphics[width=1.0\columnwidth]{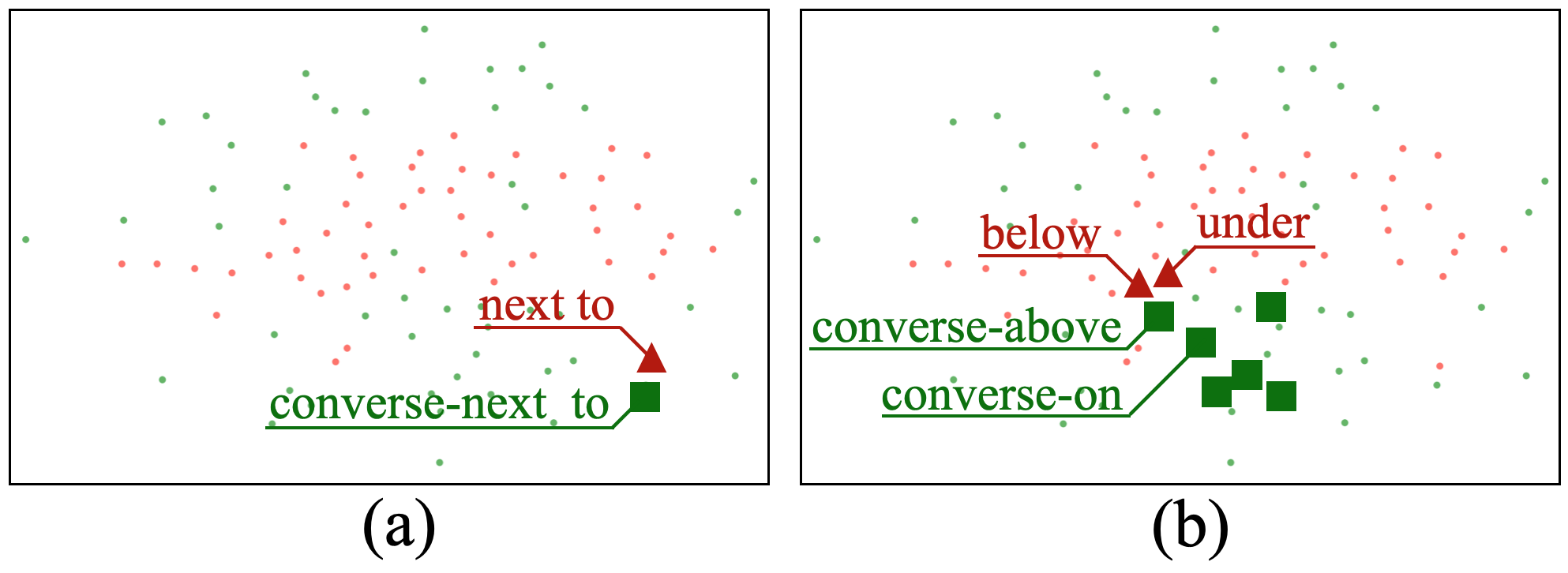}
\caption{\textbf{t-SNE visualization of relationship features $\vf^R$.} (a) With the relationship feature of \textit{next\_to} highlighted in dark red triangle, its converse version (i.e., \textit{converse-next\_to}) is its synonym (shown in dark green square). Thus, their $\vf^R$ are close to each other. (b) With \textit{below} and \textit{under} as synonyms (i.e., the two dark red triangles), the derived converse relationship features of their antonyms such as \textit{converse-on}, \textit{converse-above}, \textit{converse-be\_on}, \textit{converse-stand\_on}, \textit{converse-on\_top\_of} and \textit{converse-lay\_on} would be nearby as expected.
}
\label{figure:tsne}
\vspace{-6mm}
\end{figure}

As shown in Figure~\ref{figure:tsne}(a), the feature of \textit{next\_to} and \textit{converse-next\_to} are close to each other since the converse relationship of \textit{next\_to} is itself. In Figure~\ref{figure:tsne}(b), the features of \textit{below} and \textit{under} and the derived converse relationship features of their antonyms such as \textit{converse-on} and \textit{converse-above} are clustered as expected. 
Specifically, w/o exploiting the $E_C$, the cosine similarity scores between ``under'' and its (1) synonym ``below'' and (2) antonym ``above" are (1) 0.36 and (2) 0.15 respectively. By enforcing the converse property into our SG transformer, the associated scores (1) increase to 0.62 and (2) decrease to -0.93, which is consistent with the improved results reported in ablation studies of Table \textcolor{red}{D}. These experiments show that our design does exploit the information carried on converse relationship pairs.

\subsection{Three-level image outpainting}
\begin{figure*}[t!]
\centering
\vspace{-2mm}
    \includegraphics[width=0.9\linewidth]{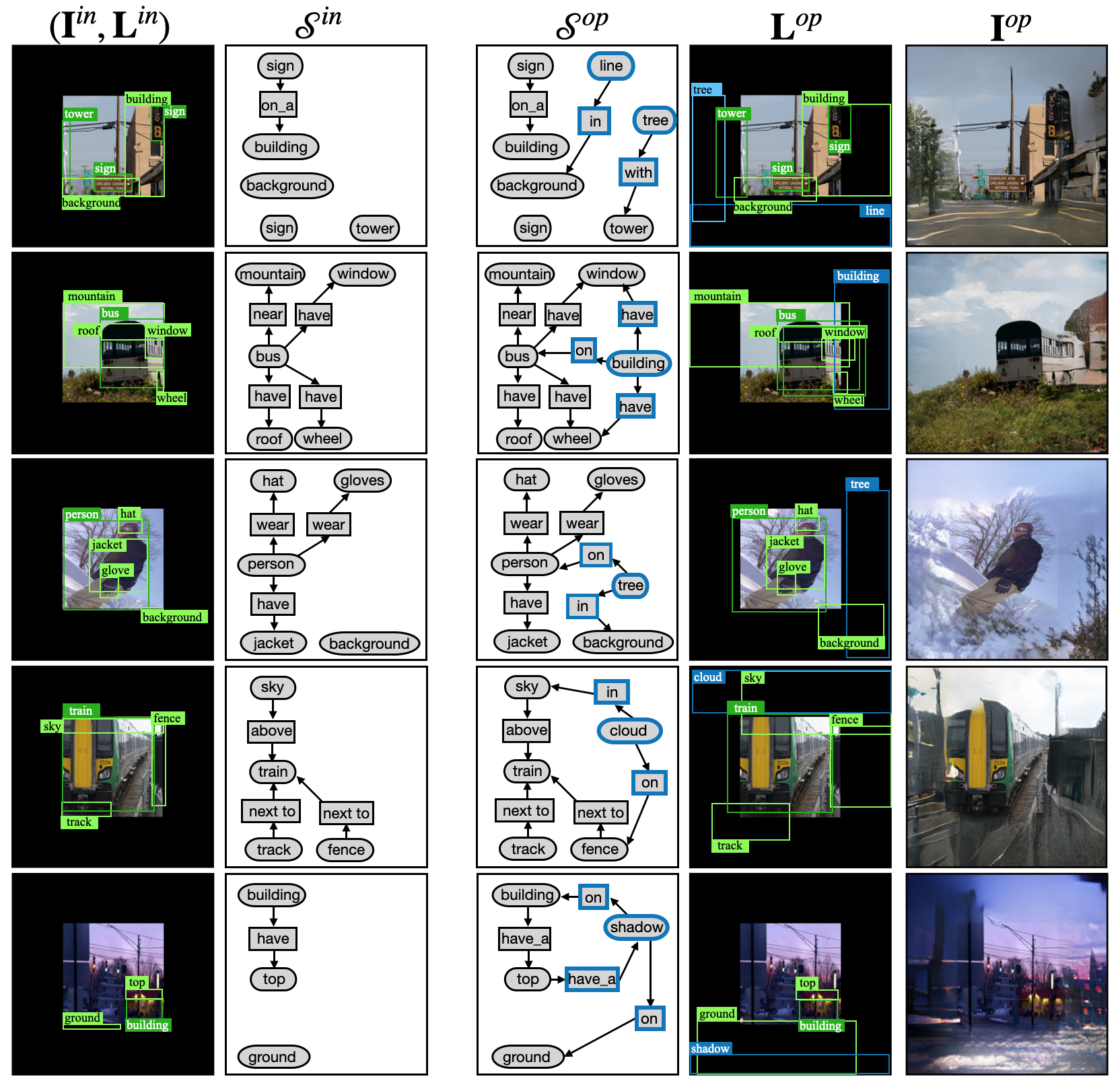}
\caption{\textbf{Visualization of our semantics-guided image outpainting on VG-MSDN.} From left to right: input image with layout $(\mI^{in}, \mL^{in})$, input scene graph $\gS^{in}$, output scene graph $\gS^{op}$, output layout $\mL^{op}$ and output image $\mI^{op}$. Note that only selected nodes from the input scene graphs are depicted for visualization purposes. Additionally, we note that when the SGE model is used for image outpainting, unlike the experiments in Figure~\red{5}(a) where only one novel object is generated, we do not enforce such constraint. That is, our SGE model is capable of generating multiple objects, e.g. the first example.
}
\label{figure:three-stage}
\end{figure*}

In Figure~\ref{figure:three-stage}, we give more examples on how our three-stage semantic image outpainting is achieved, i.e. from the extrapolation on scene graphs, to the extrapolation on layouts, then to the generation of outpainted images.

\subsection{Failure cases}
\begin{figure*}[t]
\centering
    \includegraphics[width=0.9\linewidth]{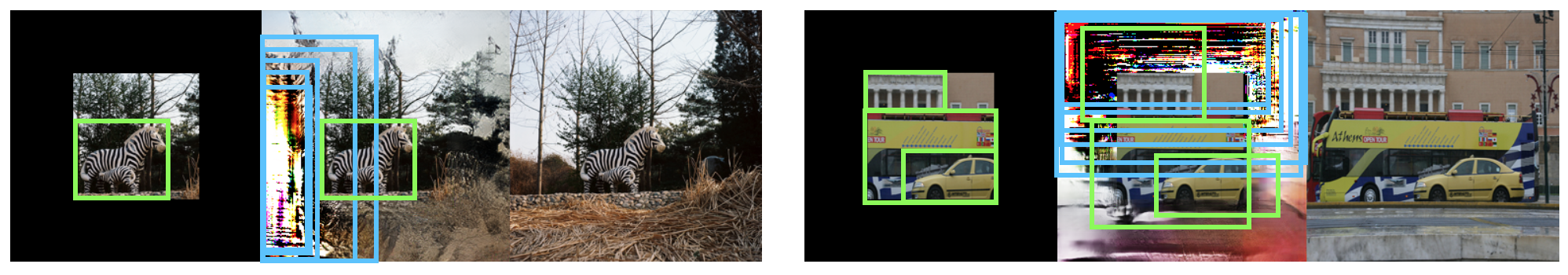}
\caption{\textbf{Two example failure cases.} From left to right: input image $\mI^{in}$,  our output image $\mI^{op}$ and ground truth image $\mI^{gt}$. Our model tends to generate images with noisy or repeating content (e.g. repeating ``legs" or ``windows" which are highlighted as blue bounding boxes) if the training images with similar visual concepts are over annotated. For example, training images of ``zibra" are often annotated with multiple (more than 4) ``legs", or those of ``building" are typically with a large number of ``windows" annotated.}
\label{figure:failure}
\vspace{-4mm}
\end{figure*}

We provide a few failure cases in Figure~\ref{figure:failure}, which are due to over-annotated data and thus prevent the learning of proper scene graphs.

\subsection{Additional Visualization}
\label{sec:visual-comparison}
\begin{figure*}[t!]
\centering
\vspace{0mm}
    \includegraphics[width=0.9\linewidth]{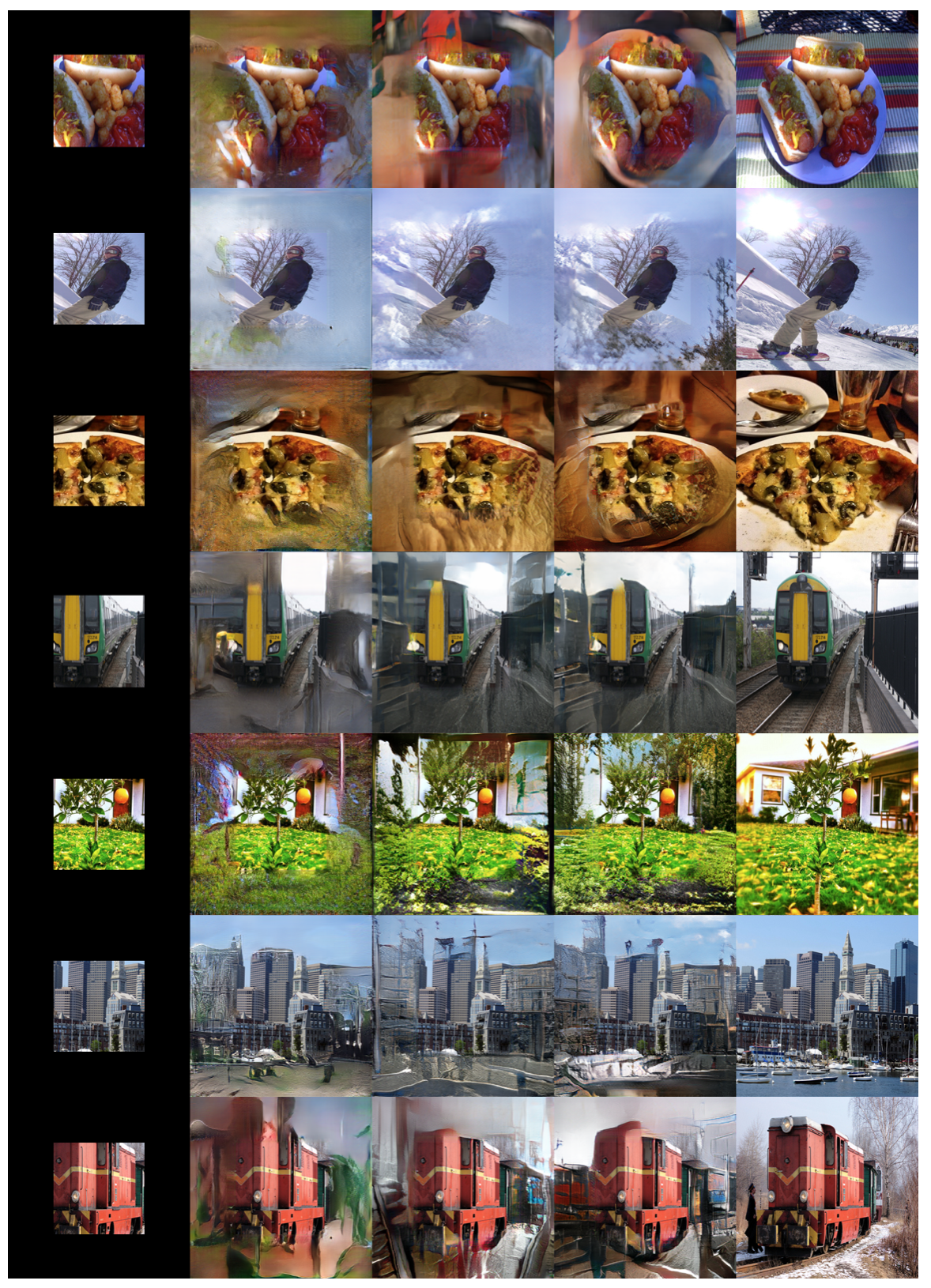}
\caption{
\textbf{Example outpainting results on VG-MSDN.}
From left to right: input image, output images produced by Boundless~\cite{teterwak2019boundless}, AttSpade~\cite{herzig2020learning} \& ours, and the ground truth image.
}
\label{figure:vis-vg-1}
\vspace{-0mm}
\end{figure*}

\begin{figure*}[t!]
\centering
\vspace{0mm}
    \includegraphics[width=0.9\linewidth]{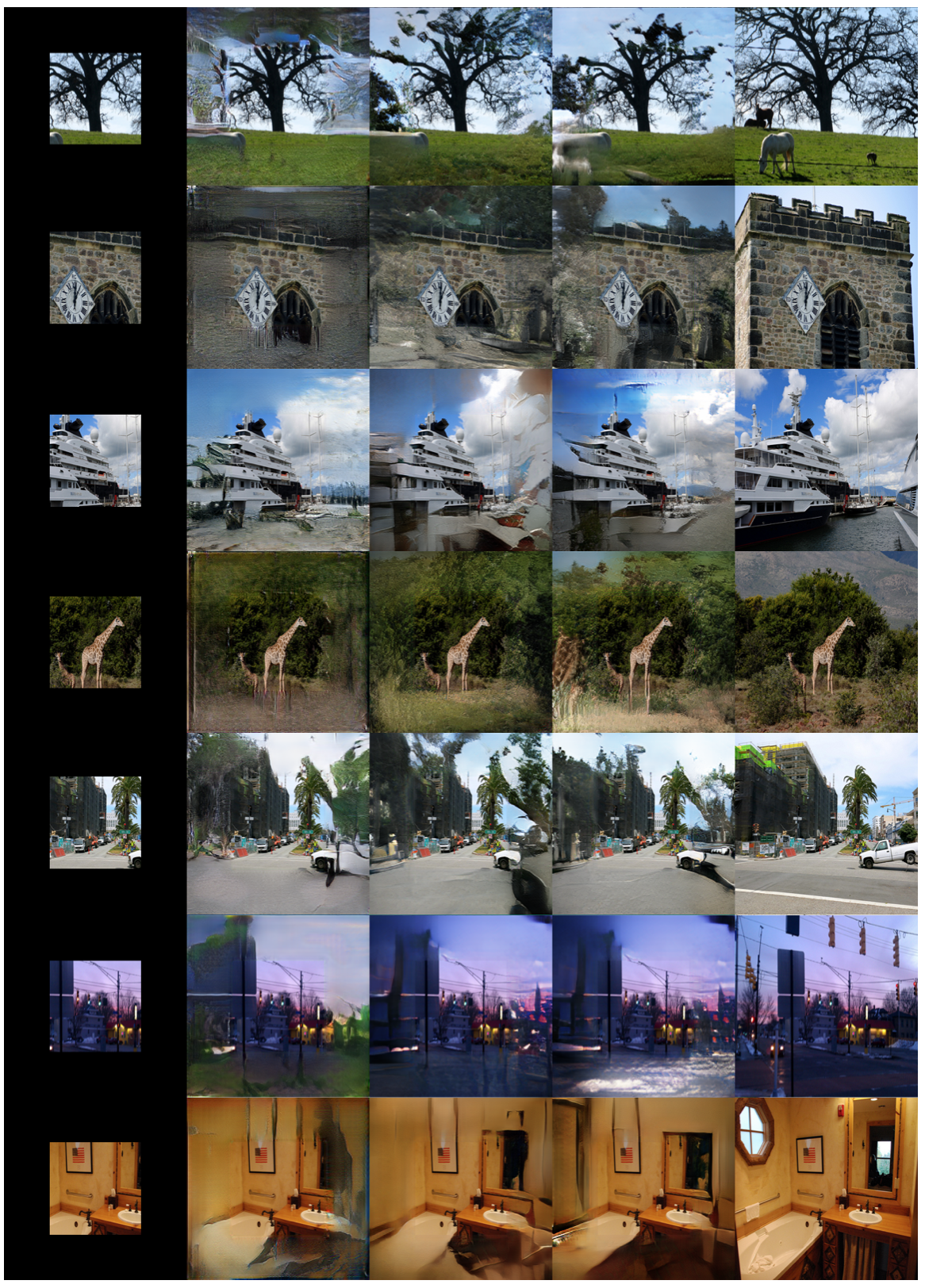}
\caption{
\textbf{Example outpainting results on VG-MSDN.}
From left to right: input image, output images produced by Boundless~\cite{teterwak2019boundless}, AttSpade~\cite{herzig2020learning} \& ours, and the ground truth image.
}
\label{figure:vis-vg-2}
\vspace{-0mm}
\end{figure*}

\begin{figure*}[t!]
\centering
\vspace{0mm}
    \includegraphics[width=0.9\linewidth]{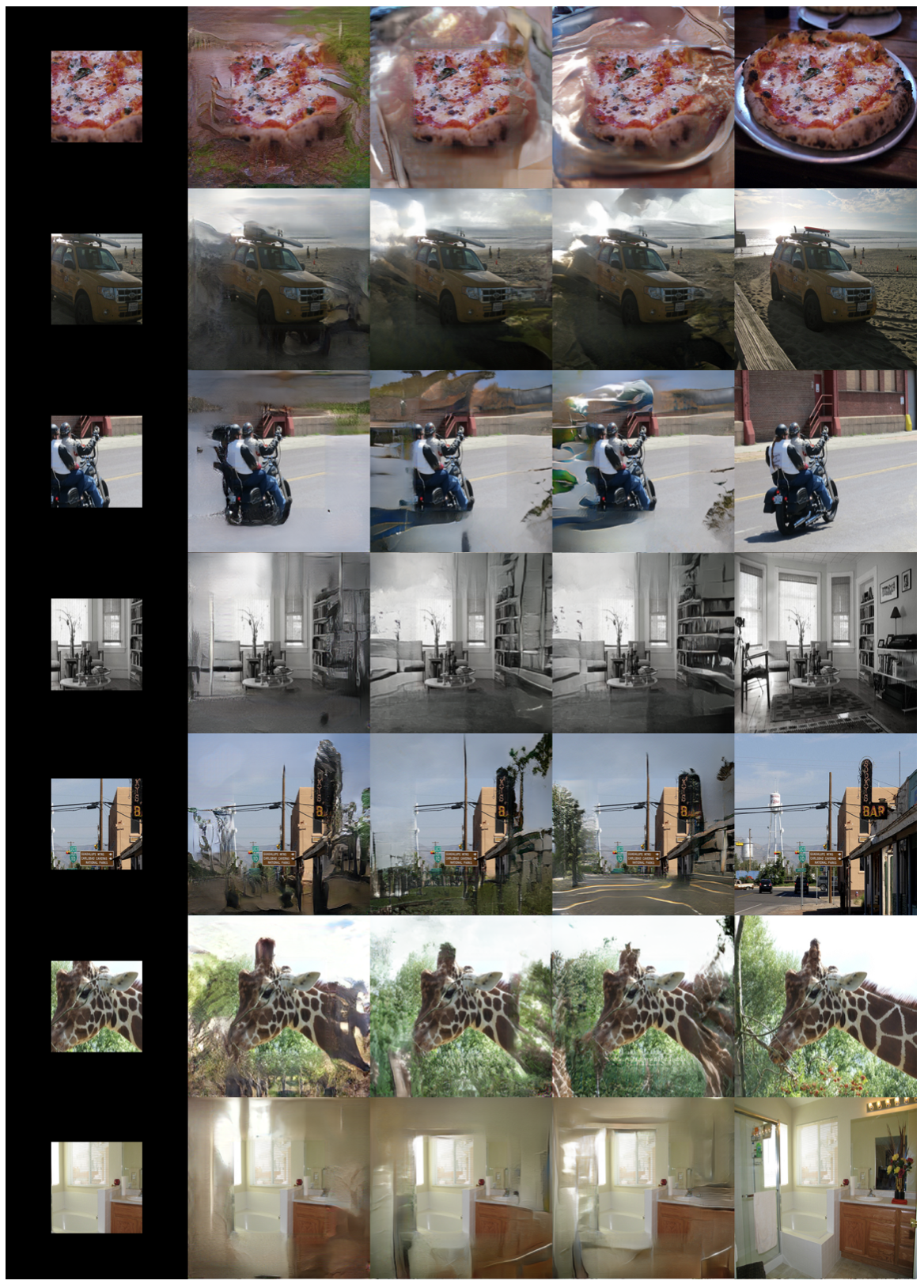}
\caption{
\textbf{Example outpainting results on VG-MSDN.}
From left to right: input image, output images produced by Boundless~\cite{teterwak2019boundless}, AttSpade~\cite{herzig2020learning} \& ours, and the ground truth image.
}
\label{figure:vis-vg-3}
\vspace{-0mm}
\end{figure*}

Additional image outpainting results on VG-MSDN are shown in Figure~\ref{figure:vis-vg-1},~\ref{figure:vis-vg-2} and ~\ref{figure:vis-vg-3}. For example, one can see the first and fourth rows of Fig. \textcolor{red}{F}, in which instances of novel/additional animal categories are introduced. In the fifth row of Fig. \textcolor{red}{G}, our model is able to synthesize the entire road region with traffic lines which are not presented in the input.

%% file: Tables/table_ablation.tex
\begin{table}[t]
\centering
\caption{Ablation studies of SGE on VG-MSDN.}
\vspace{-2.5mm}
\label{table:ablation}
\begin{tabular}{ l c c c c } 
\toprule
{} & \multicolumn{2}{c}{\textbf{Obj}} & \multicolumn{2}{c}{\textbf{Rel}} \\ 
\cmidrule(lr){2-3}
\cmidrule(lr){4-5}
{} & {rAVG} & {Hit@1 / 5} & {rAVG} & {Hit@1 / 5}  \\
\midrule
{node-level} & {9.32} & {35.7 / 64.8} & {3.69} & {46.1 / 81.6} \\
\midrule
{+ edge-level} & {8.68} & {38.7 / 68.6} & {3.80}& {48.6 / 80.7} \\
{+ $\gL_{sym}$} & {8.96} & {38.2 / 67.8} & {3.65} & {52.0 / 82.4} \\
\midrule
{Ours} & \textbf{8.38} & \textbf{39.7 / 68.9} &  \textbf{3.43} & \textbf{55.3 / 84.3}  \\
\bottomrule
\end{tabular}
\end{table}